\documentclass[journal,letterpaper,twoside]{IEEEtran}

\usepackage[utf8]{inputenc}
\usepackage[T1]{fontenc}
\usepackage{graphicx}
\DeclareGraphicsExtensions{.png,.pdf}
\usepackage{xcolor}
\usepackage[cmex10]{amsmath}
\usepackage{amsthm,amssymb,amsfonts,mathrsfs,bm,mathtools,cases}
\usepackage{subfig}
\usepackage{etoolbox}
\usepackage[linesnumbered,vlined,ruled,commentsnumbered]{algorithm2e}
\usepackage{xspace}

\makeatletter
\newcommand*{\ie}{%
  \@ifnextchar{,}%
  {{i.e.}}%
  {{i.e.,}\@\xspace}%
}
\newcommand*{\eg}{%
  \@ifnextchar{,}%
  {{e.g.}}%
  {{e.g.,}\@\xspace}%
}
\newcommand*{\etc}{%
  \@ifnextchar{.}%
  {{etc}}%
  {{etc.}\@\xspace}%
}
\newcommand*{\etal}{%
  \@ifnextchar{.}%
  {{et al}}%
  {{et al.}\@\xspace}%
}
\newcommand*{\cf}{%
  \@ifnextchar{.}%
  {{cf}}%
  {{cf.}\@\xspace}%
}
\newcommand*{\aka}{%
  \@ifnextchar{,}%
  {{a.k.a.}}%
  {{a.k.a.}\@\xspace}%
}
\makeatother

\usepackage{booktabs}
\usepackage{threeparttable}
\usepackage{multirow}
\usepackage{multicol}
\colorlet{negro}{black}
\colorlet{gris}{black!70}
\colorlet{rojo}{red!70!black}
\colorlet{rojol}{red}

\newcommand{\Integer}{\mathbb{Z}}
\newcommand{\IntegerP}{\mathbb{Z}_{\geq 0}}
\newcommand{\IntegerPP}{\mathbb{Z}_{>0}}
\newcommand{\Real}{\mathbb{R}}

\newcommand{\RealPP}{\mathbb{R}_{>0}}

\newcommand\given{{\mathbin{}\mid\mathbin{}}}
\newcommand\vect[1]{\mathbf{#1}}

\providecommand\given{} 
\newcommand\SetSymbol[1][]{
  \nonscript\,#1\vert \allowbreak \nonscript\,\mathopen{}}
\DeclarePairedDelimiterX\Set[1]{\lbrace}{\rbrace}%
{ \renewcommand\given{\SetSymbol[\delimsize]} #1 }
\DeclarePairedDelimiterX\innerp[2]{\langle}{\rangle}{#1
  \mathop{}\delimsize\vert\mathop{} #2}
\DeclarePairedDelimiterX\norm[1]\lVert\rVert{\ifblank{#1}{\:\cdot\:}{#1}}

\DeclareMathOperator{\Expect}{\mathbb{E}}
\DeclareMathOperator{\Prob}{\mathbb{P}}

\DeclareMathOperator{\rank}{rank}

\DeclareMathOperator*{\Argmin}{arg\,min}

\DeclareMathOperator{\Kprod}{\otimes_{\mathcal{H}}}

\theoremstyle{definition}
\newtheorem{prop}{Proposition}

\begin{document}

\title{Brain-Network Clustering via\\ Kernel-ARMA Modeling and the Grassmannian}

\author{%
  Cong Ye,\IEEEauthorrefmark{1} Konstantinos~Slavakis,\IEEEauthorrefmark{1},
  Pratik V.~Patil,\IEEEauthorrefmark{1}\\ Sarah~F.~Muldoon,\IEEEauthorrefmark{2}
  and John~Medaglia,\IEEEauthorrefmark{3}%
  \thanks{\IEEEauthorrefmark{1}C.~Ye, K.~Slavakis and P.~V.~Patil are with the
    Department of Electrical Engineering, University at Buffalo, The State
    University of New York (SUNY), NY 14260, USA.}%
  \thanks{\IEEEauthorrefmark{2}S.~F.~Muldoon is with the Department of
    Mathematics and the Computational and Data-Enabled Science and Engineering
    Program, University at Buffalo, SUNY, NY 14260, USA.}%
  \thanks{\IEEEauthorrefmark{3}J.~Medaglia is with the Department of Psychology,
    Drexel University, PA 19104, USA, and the Perelman School of Medicine,
    University of Pennsylvania, PA 19104, USA.}%
}

\maketitle

\begin{abstract}
  Recent advances in neuroscience and in the technology of functional magnetic
  resonance imaging (fMRI) and electro-encephalography (EEG) have propelled a
  growing interest in brain-network clustering via time-series
  analysis. Notwithstanding, most of the brain-network clustering methods
  revolve around state clustering and/or node clustering (a.k.a.\ community
  detection or topology inference) within states. This work answers first the
  need of capturing non-linear nodal dependencies by bringing forth a novel
  feature-extraction mechanism via kernel autoregressive-moving-average
  modeling. The extracted features are mapped to the Grassmann manifold
  (Grassmannian), which consists of all linear subspaces of a fixed rank. By
  virtue of the Riemannian geometry of the Grassmannian, a unifying clustering
  framework is offered to tackle all possible clustering problems in a network:
  Cluster multiple states, detect communities within states, and even
  identify/track subnetwork state sequences. The effectiveness of the proposed
  approach is underlined by extensive numerical tests on synthetic and real
  fMRI/EEG data which demonstrate that the advocated learning method compares
  favorably versus several state-of-the-art clustering schemes.
\end{abstract}

\begin{IEEEkeywords}
  Brain network, clustering, ARMA, kernel, Grassmann.
\end{IEEEkeywords}

\IEEEpeerreviewmaketitle

\section{Introduction}

Recent advances in neuroscience reveal the brain to be a complex network capable
of integrating and generating information from external and internal sources in
real time~\cite{sporns2004organization}. The rapidly growing field of Network
Neuroscience uses network analytics to reveal, via graph theory and its concepts
(\eg, nodes and edges), topological and functional dependencies of the
brain~\cite{bassett2017network}. At the microscopic scale, nodes of a brain
network correspond to individual neurons, while edges might describe synaptic
coupling between neurons or relationships between their firing
patterns~\cite{Feldt:2011fh}. At the macroscopic scale, nodes might be brain
regions, and edges might represent anatomical connections (structural
connectivity) or statistical relationships between regional brain dynamics
(functional connectivity)~\cite{Bullmore:2009iv}.

Popular noninvasive techniques used to acquire time series data from brain
networks include functional magnetic resonance imaging (fMRI) and
electro-encephalography (EEG). In particular, fMRI monitors the blood
oxygen-level dependent (BOLD) time series~\cite{ogawa1990brain}, while EEG
tracks brain activity through the time series which are collected via electrodes
on the scalp. EEG possesses a high temporal resolution and is considered to be
relatively convenient, inexpensive, and harmless compared to other methods such
as magneto-encephalography (MEG), which is much less risky than positron
emission tomography (PET)~\cite{wang2013new}.

An important aspect of the majority of works in network analytics is that the
time-series data describing the nodal signals tend to be considered stationary,
and many learning algorithms make the temporal smoothness assumption
\cite{Folino:2014dc, Mateos:SPM:19}. However, stationarity of the brain network
data should not be assumed, as it is known that the brain acts as a
non-stationary network even during its resting state, \eg
\cite{Britz.EEG.RSN.10, liu2016combined, chronnectome, allen2014tracking,
  Vaiana:2018gb}. Dynamic functional brain networks can be built using pairwise
relationships derived from the time series data described above, and this
dynamic-network viewpoint has been widely exploited to identify diseases,
cognitive states, and individual differences in
performance~\cite{hutchison2013dynamic, kucyi2014dynamic, kaiser2016dynamic,
  christoff2016mind}.


Learning algorithms are often employed to identify functional dependencies among
nodes and topology in networks. As a prominent example, clustering algorithms
have been already utilized to verify the dynamic nature of brain
networks~\cite{Britz.EEG.RSN.10, liu2016combined}, as well as to predict and
detect brain disorders, applied to syndromes at large, such as
depression~\cite{Greicius.depression.07}, epilepsy,
schizophrenia~\cite{Broyd.default.09}, Alzheimer disease and
autism~\cite{Stam.Alzheimer.09}. In general, brain-network clustering methods
aim at three major goals: Node clustering (a.k.a.\ community detection or
topology inference) within a given brain state, state clustering of similar
brain states, and subnetwork-state-sequence identification. Loosely speaking, a
``brain state'' corresponds to a specific ``global'' network topology or nodal
connectivity pattern which stays fixed over a time interval. A ``subnetwork
state sequence'' is defined as the latent (stochastic) process that drives a
subnetwork/subgroup of nodal time series, may span several
network-wide/``global'' states, and the collaborating nodes may even change as
the brain transitions from one ``global'' state to another.

Most brain-clustering algorithms are used for nodal and state clustering, while
only very few schemes try to identify/track subnetwork state sequences. For
example, community detection in brain networks has been studied extensively to
perform clustering in both static and dynamic brain networks
\cite{Garcia:2018gz, Rossetti:CDD:18}. Modularity maximization
\cite{Newman:2006tx, Mucha:2010bz} is a popular method for performing community
detection in functional brain networks, but also relies on the selection of
additional parameters determining the proper null model, resolution of
clustering and in the dynamic case, the value of interlayer coupling
\cite{Vaiana:2018gba}.  In~\cite{orhan2011eeg, orhan2012epileptic}, the discrete
wavelet transform decomposes EEG signals into frequency sub-bands, and K-means
is used to cluster the acquired wavelet coefficients. In~\cite{andersen2014non},
non-parametric Bayesian models, coined Bayesian community detection and infinite
relational modeling, are introduced and applied to resting state fMRI data to
define probabilities on network edges. Clustering in~\cite{andersen2014non} is
performed by running sophisticated comparisons on the values of those edge
probabilities. Study~\cite{martens2017brain} investigated network ``motifs,''
defined as recurrent and statistically significant sub-graphs or patterns. A
spectral-clustering based algorithm applied to motif features revealed a
spatially coherent, yet frequency dependent, sub-division between the posterior,
occipital and frontal brain regions~\cite{martens2017brain}. Entropy
maximization and frequency decomposition were utilized
in~\cite{mizuno2010clustering} prior to applying vector
quantization~\cite{sato1996generalized} to frequency-based features for
clustering communities within EEG data. In~\cite{mammone2011clustering},
EEG-data topography via Renyi's entropy was proposed as a feature extraction
mapping, before applying self-organizing maps as the off-the-shelf clustering
algorithm. In the recently popular graph-signal-processing
context~\cite{Mateos:SPM:19, Segarra:17}, topology inference is achieved by
solving optimization problems formed via the observed time-series data and the
eigen-decomposition of the Laplacian matrix of the network.

Other approaches have been used to perform state clustering.  For example,
\cite{ou2015characterizing} advocates hidden Markov models (HMMs) to
characterize brain-state dynamics. HMM parameters are extracted from each state
and used to form vectors in a Euclidean space, with their pairwise metric
distances comprising the entries of an affinity matrix. Hierarchical clustering
is then applied to the affinity matrix to cluster brain
states. In~\cite{ma2014dynamic}, time-varying features are extracted from
healthy controls and patients with schizophrenia using independent vector
analysis (IVA). Mutual information among the IVA features, Markov modeling and
K-means are used to detect changes in the brain's spatial connectivity
patterns. A change-point detection approach for resting state fMRI is introduced
in~\cite{li2013detecting}. Functional connectivity patterns of all
fiber-connected cortical voxels are concatenated into a descriptive feature
vector to represent the brain's state, and the temporal change points of
different brain states are decided by detecting the abrupt changes of the vector
patterns via a sliding window approach. In~\cite{Masuda:ScientificReports:19},
hierarchical clustering is applied to a time series of graph-distance measures
to identify discrete states of networks. Moreover, motivated by the observation
that changes in nodal communities suggest changes in network states, studies
\cite{al2017tensor, al2018tensor} perform community detection on fMRI data,
prior to state clustering, by capitalizing on K-means, multi-layer modeling,
(Tucker) tensor and higher-order singular value decomposition.

There is only a few methods that can cluster subnetwork state sequences in fMRI
and EEG modalities. In~\cite{brechet2019capturing}, features are extracted from
the frequency content of the fMRI/EEG time series. A feature example is the
ratio of the sum of amplitudes within a specific frequency interval over the sum
of amplitudes over the whole frequency range of the time series. Features, and
thus subnetwork state sequences, are then clustered via
K-means~\cite{brechet2019capturing}. A computer-vision approach is introduced
in~\cite{lim2018novel}. EEG data are transformed into dynamic topographic maps,
able to display features such as voltage amplitude, power and peak latency. The
flow of activation within those topographic maps is estimated by using an
optical-flow estimation method~\cite{horn1981determining} which generates motion
vectors. Motion vectors are clustered into groups, and these dynamic clusters
are tracked along the time axis to depict the activation flow and track the
subnetwork state sequences.


This paper capitalizes on the directions established
by~\cite{slavakis2018clustering} to introduce a \textit{unifying}\/
feature-extraction and clustering framework, with strong geometric flavor, that
make no assumptions of stationarity and can carry through all possible
brain-clustering duties, \ie, community detection, state clustering, and
subnetwork-state-sequence clustering/tracking. A kernel
autoregressive-moving-average (K-ARMA) model is proposed to capture latent
non-linear and causal dependencies, not only within a single time series, but
also among multiple nodal time series of the brain network. To accommodate the
highly likely non-stationarity of the time series, the K-ARMA model is applied
via a time-sliding window. Per application of the K-ARMA model, a system
identification problem is solved to extract a low-rank observability
matrix. Such a low-rank representation enables dimensionality-reduction
arguments which are beneficial to learning methods for the usually
high-dimensional ambient spaces associated with brain-network
analytics. Features are defined as the low-rank column spaces of the computed
observability matrices. For a fixed rank, those features become points of the
Grassmann manifold (Grassmannian), which enjoys the rich Riemannian
geometry. This feature-extraction scheme permeates all clustering duties in this
study. Having obtained the features and to identify clusters, this study builds
on Riemannian multi-manifold modeling (RMMM)~\cite{RMMM.AISTATS.15, RMMM.arxiv,
  slavakis2018clustering}, which postulates that clusters take the form of
sub-manifolds in the Grassmannian. To compute clusters, the underlying
Riemannian geometry is exploited by the geodesic-clustering-with-tangent-spaces
(GCT) algorithm~\cite{RMMM.AISTATS.15, RMMM.arxiv,
  slavakis2018clustering}. Unlike the pipeline in \cite{yger2017riemannian},
which used covariance matrix of EEG, after low-pass filtering, as feature on a
manifold and considered only the Riemannian distance between features, GCT
considers both distance and angle as geometric information for clustering. In
contrast to~\cite{RMMM.AISTATS.15, RMMM.arxiv, slavakis2018clustering}, where
the number of clusters need to be known a priori, this paper incorporates
hierarchical clustering to render GCT free from any a-priori knowledge of the
number of clusters. Extensive numerical tests on synthetic and real fMRI/EEG
data demonstrate that the proposed framework, \ie, feature extraction mechanism
and GCT-based clustering algorithm, compares favorably versus state-of-the-art
manifold learning and brain-network clustering schemes.

The rest of the paper is organized as follows. The K-ARMA model and the
feature-extraction mechanism are introduced in Section~\ref{Sec:K-ARMA}. The new
variant of the GCT clustering algorithm is presented in
Section~\ref{Sec:Clustering}, while synthetic and real fMRI/EEG data are used in
Section~\ref{Sec:Tests} to validate the theoretical and algorithmic
developments. The manuscript is concluded in Section~\ref{Sec:conclusions},
while mathematical notation, any background material as well as proofs are
deferred to the Appendix.

\section{Kernel-ARMA Modeling}\label{Sec:K-ARMA}

\begin{figure}[!t]
  \centering
  \includegraphics[width = .9\linewidth]{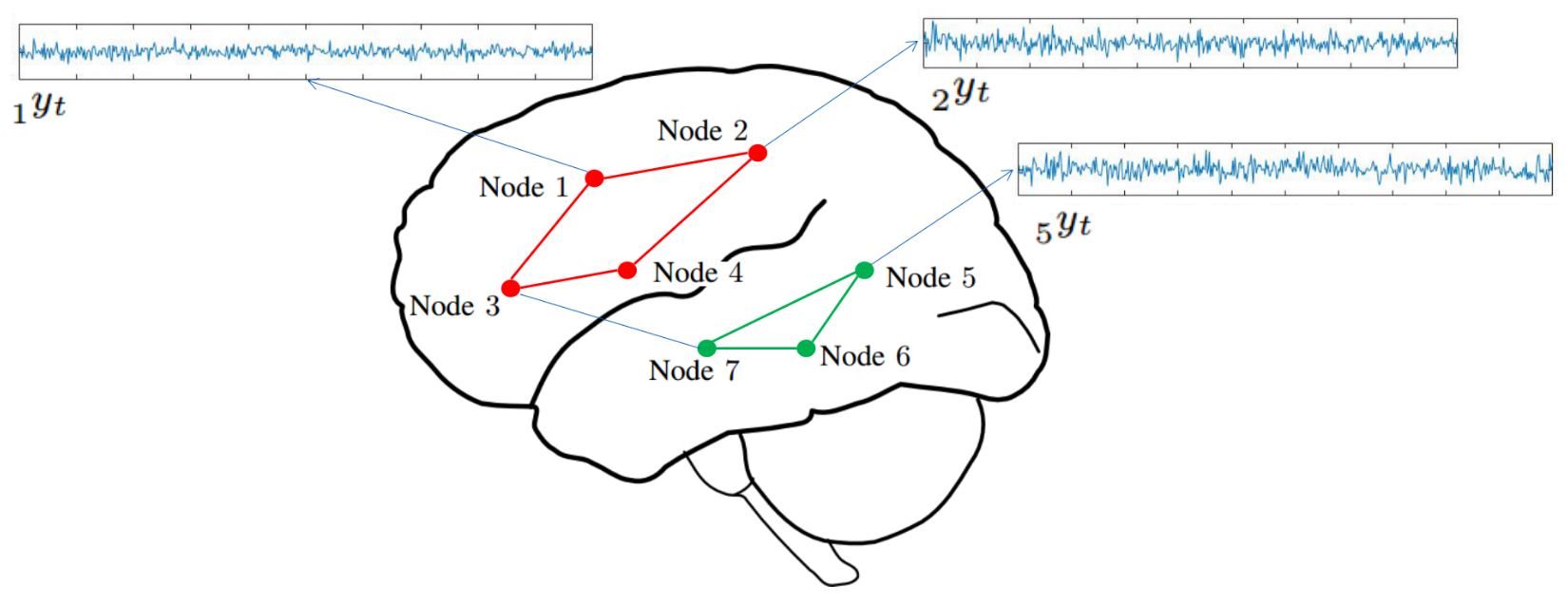}
  \caption{Brain network with nodes, edges, and nodal time series (signals)
    $\prescript{}{\nu}{y}_{t}$.}\label{Fig:brain.net}
\end{figure}

Consider a (brain) network/graph
$\mathcal{G} \coloneqq (\mathcal{N}, \mathcal{E})$, with sets of nodes
$\mathcal{N}$, of cardinality $\lvert\mathcal{N} \rvert$, and edges
$\mathcal{E}$. Each node $\nu\in \mathcal{N}$ is annotated by a discrete-time
stochastic process (time series) $(\prescript{}{\nu}{y}_{t})_{t\in \Integer}$,
where $t$ denotes discrete time and $\Integer$ the set of all integer numbers;
\cf Fig.~\ref{Fig:brain.net}. To avoid congestion in notations,
$\prescript{}{\nu}{y}_{t}$ stands for both the random variable (RV) and its
realization. The physical meaning of $\mathcal{N}$ and
$(\prescript{}{\nu}{y}_{t})_{t\in \Integer}$ depends on the underlying
data-collection modalities. For example, in fMRI, nodes $\mathcal{N}$ comprise
regions of interest (ROI) of the brain which are connected either anatomically
or functionally, and $(\prescript{}{\nu}{y}_{t})_{t\in \Integer}$ becomes a BOLD
time series of the average signal in a given ROI~\cite{ogawa1990brain}, \eg, Fig.~\ref{Fig:fMRI.BOLD}. In EEG,
$\mathcal{N}$ consists of all electrodes placed on the scalp, and
$(\prescript{}{\nu}{y}_{t})_{t\in \Integer}$ gathers the signal samples
collected by electrode $\nu$; \cf~Fig.~\ref{Fig:EEG.signal}. For index
$\mathcal{V}$ and an integer $q\in\IntegerPP$, the $q\times 1$ vector
$\prescript{}{\mathcal{V}}{\vect{y}}_{t}$ is used in this manuscript to collect
all signal samples from node(s) $\mathcal{V}$ of the network at the time
instance $t$, and to unify several scenarios of interest as the following
discussion demonstrates.

\subsection{State Clustering ($\mathcal{V} \coloneqq \mathcal{N}$)}
\label{Sec:state.clustering}

Since a ``state'' is a global attribute of the network across all nodes, vector
$\prescript{}{\mathcal{N}}{\vect{y}}_{t} \coloneqq [\prescript{}{1}{y}_t,
\ldots, \prescript{}{\lvert \mathcal{N} \rvert}{y}_t]^{\intercal}$, with
$\mathcal{V} \coloneqq \mathcal{N}$ and $q \coloneqq \lvert \mathcal{N} \rvert$,
stands as the "snapshot" of the network at the time instance $t$. Features will
be learned from the time series $(\prescript{}{\mathcal{N}}{\vect{y}}_{t})_t$ in
Sec.~\ref{Sec:KARMA.features} to monitor the evolution of the network and to
cluster states in Sec.~\ref{Sec:Clustering}.

\subsection{Community Detection and Clustering of Subnetwork State Sequences
  ($\mathcal{V} \coloneqq \nu$)} \label{Sec:comm.task.clustering}

In the case of community detection and subnetwork-state-sequence clustering,
nodes $\mathcal{N}$ need to be partitioned via the (dis)similarities of their
time series. For example, in subnetwork-state-sequence clustering, same-cluster
nodes collaborate to carry through a common task. To be able to detect common
features and to identify those nodes, it is desirable first to extract
individual features from each nodal time series. To this end, $\mathcal{V}$ is
assigned the value $\nu$, so that $\forall \nu\in\mathcal{N}$, for a given
buffer length $\texttt{Buff}_{\nu}\in \IntegerPP$ and with
$q = \texttt{Buff}_{\nu}$, $\prescript{}{\nu}{\vect{y}}_{t}$ takes the form of
$[\prescript{}{\nu}{y}_t, \ldots, \prescript{}{\nu}{y}_{t+ \texttt{Buff}_{\nu}
  -1}]^{\intercal}$.

\subsection{Extracting Grassmannian Features}\label{Sec:KARMA.features}

Consider now a user-defined RKHS $\mathcal{H}$ with its kernel mapping
$\varphi(\cdot)$; \cf~App.~\ref{App:RKHS}. Given $N\in\IntegerPP$ and
assuming that the sequence $(\prescript{}{\mathcal{V}}{\vect{y}}_{t})_t$ is
available, define
$\bm{\varphi}_t \coloneqq [\varphi(\prescript{}{\mathcal{V}}{\vect{y}}_{t}),
\varphi(\prescript{}{\mathcal{V}}{\vect{y}}_{t+1}), \ldots,
\varphi(\prescript{}{\mathcal{V}}{\vect{y}}_{t+N-1})]^{\intercal} \in
\mathcal{H}^N$. This work proposes the following kernel (K-)ARMA model to fit
the variations of features $\Set{\bm{\varphi}_t}_t$ within space $\mathcal{H}$:
There exist matrices $\vect{C}\in \Real^{N\times\rho}$,
$\vect{A}\in \Real^{\rho\times \rho}$, the latent variable
$\bm{\psi}_t\in \mathcal{H}^{\rho}$, and vectors
$\bm{\upsilon}_t\in \mathcal{H}^N$, $\bm{\omega}_t\in \mathcal{H}^{\rho}$ that
capture noise and approximation errors, s.t.\ $\forall t$,
\begin{subequations}\label{KARMA}
  \begin{align}
    \bm{\varphi}_t & = \vect{C} \bm{\psi}_t + \bm{\upsilon}_t \,, \\
    \bm{\psi}_t & = \mathbf{A} \bm{\psi}_{t-1} + \bm{\omega}_t \,.
  \end{align}
\end{subequations}

\begin{prop}\label{Prop:O}
  Given parameter $m\in\IntegerPP$, define the ``forward'' matrix-valued
  function
  \begin{subequations}\label{FB} 
    \begin{align}%
      \bm{\mathcal{F}}_t
      & \coloneqq \begin{bmatrix}
        \bm{\varphi}_t & \bm{\varphi}_{t+1} & \ldots &
        \bm{\varphi}_{t+\tau_{\text{f}}-1} \\ 
        \bm{\varphi}_{t+1} & \bm{\varphi}_{t+2} & \ldots &
        \bm{\varphi}_{t+\tau_{\text{f}}} \\ 
        \vdots & \vdots & \ddots & \vdots \\ 
        \bm{\varphi}_{t+m-1} & \bm{\varphi}_{t+m} & \ldots &
        \bm{\varphi}_{t+\tau_{\text{f}}+m-2} 
      \end{bmatrix} \in \mathcal{H}^{mN \times \tau_{\text{f}}} \,,
    \end{align}
    and the ``backward'' matrix-valued function
    \begin{align}
      \bm{\mathcal{B}}_t
      & \coloneqq \begin{bmatrix}
        \bm{\varphi}_t & \bm{\varphi}_{t+1} & \ldots &
        \bm{\varphi}_{t+\tau_{\text{f}}-1} \\ 
        \bm{\varphi}_{t-1} & \bm{\varphi}_{t} & \ldots &
        \bm{\varphi}_{t+\tau_{\text{f}}-2} \\ 
        \vdots & \vdots & \ddots & \vdots \\ 
        \bm{\varphi}_{t-\tau_{\text{b}}+1} & \bm{\varphi}_{t-\tau_{\text{b}}+2}
        & \ldots & \bm{\varphi}_{t+\tau_{\text{f}}-\tau_{\text{b}}}
      \end{bmatrix} \in\mathcal{H}^{\tau_{\text{b}}N\times \tau_{\text{f}}} \,.
    \end{align}%
  \end{subequations}%
  Then, there exist matrices
  $\bm{\Pi}_{t+1}\in \Real^{\rho\times \tau_{\text{b}}N}$ and
  $\bm{\mathcal{E}}_{t+1}^{\tau_{\text{f}}}\in \Real^{mN\times
    \tau_{\text{b}}N}$ s.t.\ the following \textit{low-rank}\/ factorization
  holds true:
  \begin{align}
    \tfrac{1}{\tau_{\text{f}}} \bm{\mathcal{F}}_{t+1} \Kprod
    \bm{\mathcal{B}}_t^{\intercal} = \vect{O} \bm{\Pi}_{t+1} +
    \bm{\mathcal{E}}_{t+1}^{\tau_{\text{f}}} \,,\label{low.rank.formula}
  \end{align}
  where product $\Kprod$ is defined in App.~\ref{App:RKHS}, and $\vect{O}$
  is the so-called 
  \textit{observability}\/ matrix:
  \begin{align*}
    \vect{O} \coloneqq \left[ \vect{C}^{\intercal}, (\vect{CA})^{\intercal}, \ldots,
    (\vect{CA}^{m-1})^{\intercal} \right]^{\intercal} \in \Real^{mN\times\rho} \,.
  \end{align*}

  With regards to a probability space, if $(\bm{\upsilon}_t)_t$ and
  $(\bm{\omega}_t)_t$ in \eqref{KARMA} are considered to be zero-mean and independent and identically distributed stochastic processes, independent of each other, if $(\bm{\omega}_t)_t$ is
  independent of $(\bm{\psi}_t)_t$, and independency holds true also between
  $(\bm{\omega}_t, \bm{\psi}_t)$, $\forall (t, t')$ s.t.\ $t > t'$, then
  \begin{align}
    \Expect\Set*{\tfrac{1}{\tau_{\text{f}}} \bm{\mathcal{F}}_{t+1} \Kprod
    \bm{\mathcal{B}}_t^{\intercal} \given \{ \bm{\psi}_{t'} \}_{t'=t-
    \tau_{\text{b}}+1}^{t+\tau_{\text{f}} + m -1} }  = \vect{O} \bm{\Pi}_{t+1}
    \,. \label{cond.expectation} 
  \end{align}

  If, in addition, $(\bm{\omega}_t)_t$, $(\bm{\upsilon}_t)_t$,
  $(\bm{\psi}_t)_t$, and
  $(\bm{\omega}_t \Kprod \bm{\psi}_{t-\tau}^{\intercal})_t$,
  $\forall \tau\in\IntegerPP$, are wide-sense stationary, then
  $\lim_{\tau_{\text{f}} \to \infty} \bm{\mathcal{E}}_t^{\tau_{\text{f}}} =
  \vect{0}$, $\forall t$, in the mean-square ($\mathcal{L}_2$-) sense w.r.t.\
  the probability space.
  
\end{prop}

\begin{IEEEproof}
  See App.~\ref{App:prove.prop}.
\end{IEEEproof}

There can be many choices for the reproducing kernel function
$\kappa(\cdot, \cdot)$ (\cf App.~\ref{App:RKHS}). If the linear kernel
$\kappa_{\text{lin}}$ is chosen, then $\mathcal{H} = \Real^q$, $\varphi(\cdot)$
becomes the identity mapping,
$\bm{\varphi}_t = [\vect{y}_t^{\intercal}, \vect{y}_{t+1}^{\intercal}, \ldots,
\vect{y}_{t+N-1}^{\intercal}]^{\intercal} \in \Real^{qN}$, and $\Kprod$ boils
down to the usual matrix product. This case was introduced
in~\cite{slavakis2018clustering}. The most popular choice for $\kappa$ is the
Gaussian kernel $\kappa_{\text{G}; \sigma}$, where parameter $\sigma > 0$ stands
for standard deviation. However, pinpointing the appropriate $\sigma_*$ for a
specific dataset is a difficult task which may entail cumbersome
cross-validation procedures~\cite{Scholkopf.Smola.Book}. A popular approach to
circumvent the judicious selection of $\sigma_*$ is to use a dictionary of
parameters $\Set{\sigma_j}_{j=1}^J$, with $J\in\IntegerPP$, to cover an interval
where $\sigma_*$ is known to belong to. A reproducing kernel function
$\kappa(\cdot, \cdot)$ can be then defined as the convex combination
$\kappa(\cdot, \cdot) \coloneqq \sum_{j=1}^J \gamma_j \kappa_{\text{G};
  \sigma_j}(\cdot, \cdot)$, where $\Set{\gamma_j}_{j=1}^J$ are convex weights,
\ie, non-negative real numbers s.t.\
$\sum_{j=1}^J \gamma_j = 1$~\cite{Scholkopf.Smola.Book}. Such a strategy is
followed in Section~\ref{Sec:Tests}. Examples of non-Gaussian kernels
can be also found in App.~\ref{App:RKHS}.

Kernel-based ARMA models have been already studied in the context of
support-vector regression~\cite{Drezet.98, martinez2006support,
  shpigelman2009kernel}. However, those models are different than \eqref{KARMA}
since only the AR and MA vectors of coefficients are mapped to an RKHS feature
space, while the observed data $\prescript{}{\nu}{y}_{t}$ (of only a single time
series) are kept in the input space. Here, \eqref{KARMA} offers a way to map
even the observed data to an RKHS to capture non-linearities in data via
applying the ARMA idea to properly chosen feature spaces. In a different
context~\cite{Pincombe:ASOR}, time series of graph-distance metrics are fitted
by ARMA modeling to detect anomalies and thus identify states in
networks. Neither Riemannian geometry nor kernel functions were investigated
in~\cite{Pincombe:ASOR}.

Motivated by~\eqref{low.rank.formula}, \eqref{cond.expectation}, the result
($\lim_{\tau_{\text{f}} \to \infty} \bm{\mathcal{E}}_t^{\tau_{\text{f}}} =
\vect{0}$, $\forall t$), and the fact that the conditional expectation is the
least-squares-best estimator~\cite[\S9.4]{Williams.book}, the following task is
proposed to obtain an estimate of the observability matrix:
\begin{align}
  \left(\prescript{}{\mathcal{V}}{\hat{\vect{O}}}_t, \hat{\bm{\Pi}}_t \right) \in
  \Argmin_{\substack{\vect{O}\in\Real^{mN \times \rho}\\
  \bm{\Pi}\in\Real^{\rho\times \tau_{\text{b}} N}}} \norm*{\tfrac{1}{\tau_{\text{f}}}
  \bm{\mathcal{F}}_{t+1} \Kprod \bm{\mathcal{B}}_t^{\intercal} - \vect{O}
  \bm{\Pi}}_{\text{F}}^2 \,. \label{estimate.O}
\end{align}
To solve \eqref{estimate.O}, the singular value decomposition (SVD) is applied
to obtain
$(1/\tau_{\text{f}}) \bm{\mathcal{F}}_{t+1} \Kprod
\bm{\mathcal{B}}_t^{\intercal} = \vect{U} \bm{\Sigma}\vect{V}^{\intercal}$,
where $\vect{U}\in\mathbb{R}^{mN\times mN}$ is orthogonal. Assuming that
$\rho \leq \rank [(1/\tau_{\text{f}}) \bm{\mathcal{F}}_{t+1} \Kprod
\bm{\mathcal{B}}_t^{\intercal}]$, the Schmidt-Mirsky-Eckart-Young
theorem~\cite{ben2003generalized} provides the estimates
$\prescript{}{\mathcal{V}}{\hat{\vect{O}}}_t \coloneqq \vect{U}_{:,1:\rho}$ and
$\hat{\bm{\Pi}}_t \coloneqq \bm{\Sigma}_{1:\rho, 1:\rho}
\vect{V}^{\intercal}_{:,1:\rho}$, where $\vect{U}_{:,1:\rho}$ is the orthogonal
matrix that collects those columns of $\vect{U}$ that correspond to the top
(principal) $\rho$ singular values in $\bm{\Sigma}$.

Due to the factorization $\vect{O} \bm{\Pi}$, identifying the observability
matrix becomes ambiguous, since for any non-singular matrix
$\vect{P}\in \Real^{\rho\times \rho}$,
$\vect{O} \bm{\Pi} = \vect{O}\vect{P} \cdot \vect{P}^{-1} \bm{\Pi}$, and
$\prescript{}{\mathcal{V}}{\hat{\vect{O}}}_t \vect{P}$ can serve also as an
estimate. By virtue of the elementary observation that the column (range) spaces
of $\prescript{}{\mathcal{V}}{\hat{\vect{O}}}_t \vect{P}$ and
$\prescript{}{\mathcal{V}}{\hat{\vect{O}}}_t$ coincide, it becomes preferable to
identify the column space of $\prescript{}{\mathcal{V}}{\hat{\vect{O}}}_t$,
denoted hereafter by $[\prescript{}{\mathcal{V}}{\hat{\vect{O}}}_t]$, rather
than the matrix $\prescript{}{\mathcal{V}}{\hat{\vect{O}}}_t$ itself. If
$\rho = \rank [\prescript{}{\mathcal{V}}{\hat{\vect{O}}}_t]$, then
$[\prescript{}{\mathcal{V}}{\hat{\vect{O}}}_t]$ becomes a point in the Grassmann
manifold $\text{Gr}(\rho, mN)$, or Grassmannian, which is defined as the
collection of all linear subspaces of $\Real^{mN}$ with rank equal to
$\rho$~\cite[p.~73]{Loring.Tu.book}. The Grassmannian $\text{Gr}(\rho, mN)$ is a
Riemannian manifold with dimension equal to
$\rho(mN-\rho)$~\cite[p.~74]{Loring.Tu.book}. The algorithmic procedure of
extracting the feature $[\prescript{}{\mathcal{V}}{\hat{\vect{O}}}_t]$ from the
available data is summarized in Alg.~\ref{Algo:Map.O.to.Grassmannian}. To keep
notation as general as possible, instead of using all of the signal samples, a
subset $\mathfrak{T}\subset \Integer$ is considered and signal samples are
gathered in $(\prescript{}{\nu}{y}_{t})_{t\in \mathfrak{T}}$ per node $\nu$. All
generated features are gathered in step~\ref{Step:gather.all.features} of
Alg.~\ref{Algo:Map.O.to.Grassmannian}, denoted by
$\Set{x_i}_{i\in\mathfrak{I}}$, and indexed by the set $\mathfrak{I}$ of
cardinality $\lvert\mathfrak{I} \rvert$.

\begin{algorithm}[!t]
  \DontPrintSemicolon
  \SetKwInOut{input}{Input}
  \SetKwInOut{parameters}{Parameters}
  \SetKwInOut{output}{Output}
  \SetKwBlock{initial}{Initialization}{}

  \input{Time series $\Set{(\prescript{}{\nu}{y}_{t})_{t\in \mathfrak{T}}}_{\nu\in
      \mathcal{N}}$.}
  
  \parameters{Positive integers $N$, $m$, $\rho$, $\tau_{\text{f}}$ and
    $\tau_{\text{b}}$.}
  
  \output{Grassmannian features $\Set{x_i}_{i\in\mathfrak{I}}$.}

  \BlankLine
  
  Form data
  $\Set{(\prescript{}{\mathcal{V}}{\vect{y}}_t)_{t\in\mathfrak{T}}}_{\mathcal{V}\in
    \mathfrak{N}}$, where $(\mathcal{V}, \mathfrak{N})$ becomes either
  $(\mathcal{N}, \{\mathcal{N}\})$ [state clustering;
  Sec.~\ref{Sec:state.clustering}] or $(\nu, \mathcal{N})$ [community detection
  or subnetwork-state-sequence clustering; Sec.~\ref{Sec:comm.task.clustering}].

  \For{all $\mathcal{V}\in \mathfrak{N}$}{
    
    \For{all $t\in\mathfrak{T}$}{%

      Form
      $(1/\tau_{\text{f}}) \bm{\mathcal{F}}_{t+1} \Kprod
      \bm{\mathcal{B}}_t^{\intercal}$ via \eqref{FB}.

      Apply SVD:
      $(1/\tau_{\text{f}}) \bm{\mathcal{F}}_{t+1} \Kprod
      \bm{\mathcal{B}}_t^{\intercal} = \vect{U}
      \bm{\Sigma}\vect{V}^{\intercal}$.

      Feature
      $[\prescript{}{\mathcal{V}}{\hat{\vect{O}}}_t] \in \text{Gr}(\rho, mN)$ is
      the linear subspace spanned by the $\rho$ ``principal'' columns of
      $\vect{U}$.

    }
  }

  Gather all features in
  $\Set{x_i}_{i\in\mathfrak{I}} \coloneqq \cup_{\mathcal{V}\in \mathfrak{N}}
  \cup_{t\in \mathfrak{T}}
  [\prescript{}{\mathcal{V}}{\hat{\vect{O}}}_t]$.\label{Step:gather.all.features}

  \caption{Extracting Grassmannian features}\label{Algo:Map.O.to.Grassmannian}
\end{algorithm}

\section{Clustering Grassmannian Features}\label{Sec:Clustering}

\begin{figure}[!t]
  \subfloat[Clusters on $\text{Gr}(\rho,
  mN)$]{\includegraphics[width=.4\linewidth]{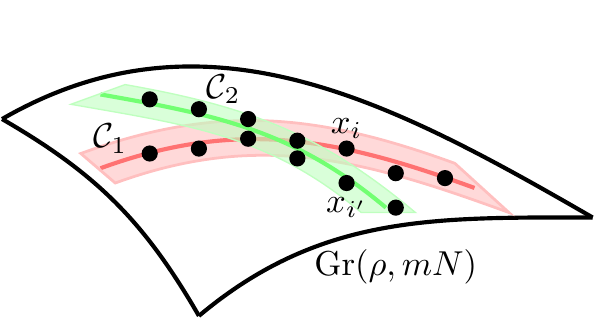}\label{Fig:RMMM}}
  \subfloat[Angular information]{\includegraphics[width =
    .6\linewidth]{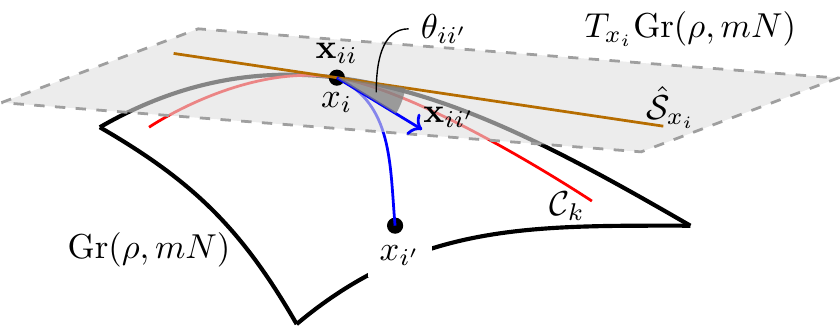}\label{Fig:theta}} 
  \caption{(a) The Riemannian multi-manifold modeling (RMMM) hypothesis. (b)
    Angular information computed in tangent spaces and used in
    Alg.~\ref{Algo:OnGrass}.}\label{Fig:RMMM.theta}
\end{figure}

Having features $\Set{x_i}_{i\in\mathfrak{I}}$ available in the Grassmannian via
Alg.~\ref{Algo:Map.O.to.Grassmannian}, the next task in the pipeline is to
cluster $\Set{x_i}_i$. This work follows the Riemannian multi-manifold modeling
(RMMM) hypothesis~\cite{RMMM.arxiv, RMMM.AISTATS.15, slavakis2018clustering},
where clusters $\Set{\mathcal{C}_k}_{k=1}^K$ are considered to be submanifolds
of the Grassmannian, with data $\Set{x_i}_i$ located close to or onto
$\Set{\mathcal{C}_k}_{k=1}^K$ (see Fig.~\ref{Fig:RMMM} for the case of $K=2$
clusters). RMMM allows for clusters to intersect; a case where the classical
K-means, for example, is known to face difficulties \cite{yang2011multi}.

Clustering is performed by Alg.~\ref{Algo:OnGrass}, coined geodesic clustering
by tangent spaces (GCT). The GCT of Alg.~\ref{Algo:OnGrass} extends its initial
form in \cite{RMMM.arxiv, RMMM.AISTATS.15, slavakis2018clustering}, since
Alg.~\ref{Algo:OnGrass} operates \textit{without}\/ the need to know the number
$K$ of clusters a-priori, as opposed to \cite{RMMM.arxiv, RMMM.AISTATS.15,
  slavakis2018clustering} where $K$ needs to be provided as input to the
clustering algorithm. This desirable feature of Alg.~\ref{Algo:OnGrass} is also
along the lines of usual practice, where it is unrealistic to know $K$ before
employing a clustering algorithm.

In a nutshell, Alg.~\ref{Algo:OnGrass} computes the affinity matrix $\vect{W}$
of features $\Set{x_i}_{i\in\mathfrak{I}}$ in
step~\ref{Step:OnGrass:adjacency.matrix}, comprising information about sparse
data approximations, via weights $\Set{\alpha_{ii'}}_{i, i'\in\mathfrak{I}}$, as
well as angles $\Set{\theta_{ii'}}_{i, i'\in\mathfrak{I}}$ between linear
subspaces. Although the incorporation of sparse weights originates from
\cite{elhamifar2011sparse}, one of the novelties of GCT is the usage of the
angular information via $\Set{\theta_{ii'}}_{i, i'\in\mathfrak{I}}$. GCT's
version of~\cite{RMMM.arxiv, RMMM.AISTATS.15, slavakis2018clustering} applies
spectral clustering in step~\ref{Step:OnGrass:Louvain}, where knowledge of the
number of clusters $K$ is necessary. To surmount the obstacle of knowing $K$
beforehand, Louvain clustering method~\cite{aynaud2010static} is adopted in
step~\ref{Step:OnGrass:Louvain}. Louvain method belongs to the family of
hierarchical-clustering algorithms that attempt to maximize a modularity
function, which monitors the intra- and inter-cluster density of
links/edges. Needless to say that any other hierarchical-clustering scheme can
be used at step~\ref{Step:OnGrass:Louvain} instead of Louvain method.

\begin{algorithm}[!t]
  \DontPrintSemicolon
  \SetKwInOut{input}{Input}
  \SetKwInOut{parameters}{Parameters}
  \SetKwInOut{output}{Output}
  \SetKwBlock{initial}{Initialization}{}

  \input{Grassmannian features $\Set{x_i}_{i\in \mathfrak{I}}$.}
  
  \parameters{$K_{\text{NN}}\in \IntegerPP$ and
    $\sigma_{\alpha}, \sigma_{\theta}\in \RealPP$.}
  
  \output{Clusters $\Set{\mathcal{C}_k}_{k=1}^K$.}

  \BlankLine

  \For{all $i\in\mathfrak{I}$\label{Step:OnGrass:every.datum}}
  {%

    Define the $K_{\text{NN}}$-nearest-neighbors
    $\mathscr{N}_{\text{NN}}(x_i)$. \label{Step:OnGrass:knn}
    
    Map $\mathscr{N}_{\text{NN}}(x_i)$ into the tangent space
    $T_{x_i}\text{Gr}(\rho, mN)$ of the Grassmannian at $x_i$ via the logarithm
    map: $\vect{x}_{ii'}\coloneqq \log_{x_i}(x_{i'})$,
    $\forall x_{i'}\in \mathscr{N}_{\text{NN}}(x_i)$. \label{Step:OnGrass:log}

    Identify $\Set{\alpha_{ii'}}_{x_{i'}\in \mathscr{N}_{\text{NN}}(x_i)}$ via
    \eqref{sparse.coding}. Set $\alpha_{ii'} \coloneqq 0$, for all $i'$ s.t.\
    $x_{i'}\notin \mathscr{N}_{\text{NN}}(x_i)$.\label{Step:OnGrass:sparse.coding}

    Compute the sample correlation matrix
    $\hat{\vect{C}}_{x_i}$ in
    \eqref{sample.covariance.matrix}. \label{Step:OnGrass:sample.covariance.matrix} 

    Perform principal component analysis (PCA) on $\hat{\vect{C}}_{x_i}$ to
    extract the eigenspace $\hat{\mathcal{S}}_{x_i}$. \label{Step:OnGrass:PCA}

    Compute angle $\theta_{ii'}$ between vector $\vect{x}_{ii'} - \vect{x}_{ii}$
    and $\hat{\mathcal{S}}_{x_i}$, $\forall x_{i'}\in \mathscr{N}_{\text{NN}}(x_i)$
    (${\theta_{ii}} \coloneqq 0$). Let also $\theta_{ii'} \coloneqq 0$ for
    $x_{i'}\notin \mathscr{N}_{\text{NN}}(x_i)$. \label{Step:OnGrass:theta}

  }

  Form the symmetric
  $\lvert \mathfrak{I}\rvert \times \lvert \mathfrak{I}\rvert$ affinity
  (adjacency) matrix $\vect{W} \coloneqq [w_{ii'}]_{(i,i')\in \mathfrak{I}^2}$,
  where entry $w_{ii'}$ is defined as \label{Step:OnGrass:adjacency.matrix}
  \[
    w_{ii'} \coloneqq \exp(\lvert \alpha_{ii'}\rvert + \lvert \alpha_{i'i}\rvert) \cdot
    \exp[-(\theta_{ii'} + \theta_{i'i})/\sigma_{\theta}] \,.
  \] 

  Apply Louvain method~\cite{aynaud2010static} to $\vect{W}$ to map the data
  $(x_i)_{i\in \mathfrak{I}}$ to clusters
  $\Set{\mathcal{C}_k}_{k=1}^K$. \label{Step:OnGrass:Louvain}

  \caption{Geodesic clustering by tangent spaces (GCT)}\label{Algo:OnGrass}
\end{algorithm}

A short description of the steps in Alg.~\ref{Algo:OnGrass} follows, with
Riemannian-geometry details deferred to \cite{RMMM.arxiv, RMMM.AISTATS.15,
  slavakis2018clustering}. Alg.~\ref{Algo:OnGrass} visits
$\Set{x_i}_{i\in\mathfrak{I}}$ sequentially
(step~\ref{Step:OnGrass:every.datum}). At step~\ref{Step:OnGrass:knn}, the
$K_{\text{NN}}$-nearest-neighbors $\mathscr{N}_{\text{NN}}(x_i)$ of $x_i$ are
identified, \ie, those $K_{\text{NN}}$ points, taken from $\Set{x_i}_i$, which
are placed the closest from $x_i$ with respect to the Grassmannian
distance~\cite{Absil.Gr.04}. The neighbors $\mathscr{N}_{\text{NN}}(x_i)$ are
then mapped at step~\ref{Step:OnGrass:log} to the Euclidean vectors
$\Set{\vect{x}_{ii'}}_{x_{i'}\in \mathscr{N}_{\text{NN}}(x_i)}$ in the tangent
space $T_{x_i}\text{Gr}(\rho, mN)$ of the Grassmannian at $x_i$ (the
gray-colored plane in Fig.~\ref{Fig:theta}) via the logarithm map
$\log_{x_i}(\cdot)$, whose computation (non-closed form via SVD) is provided in
\cite{RMMM.arxiv, slavakis2018clustering}. Step~\ref{Step:OnGrass:sparse.coding}
computes the weights
$\Set{\alpha_{ii'}}_{x_{i'} \in \mathscr{N}_{\text{NN}}(x_i)}$, with
$\alpha_{ii} \coloneqq 0$, via the following sparse-coding task:
\begin{align}
  \min_{\Set{\alpha_{ii'}}}
  {} & {} \norm*{\mathbf{x}_{ii}-\sum\nolimits_{x_{i'} \in \mathscr{N}_{\text{NN}}(x_i) \setminus \Set{x_i}}
       \alpha_{ii'}\mathbf{x}_{ii'}}^2 \notag\\  
  {} & {} + \sum\nolimits_{x_{i'} \in \mathscr{N}_{\text{NN}}(x_i) \setminus \Set{x_i}}
       \exp[{\norm{\mathbf{x}_{ii'}-\mathbf{x}_{ii}}/ \sigma_{\alpha}}]
       \cdot |\alpha_{ii'}| \notag\\ 
  \text{s.to}\ {}
     & {} \sum\nolimits_{x_{i'} \in \mathscr{N}_{\text{NN}}(x_i) \setminus \Set{x_i}}
       \alpha_{ii'} = 1 \,. \label{sparse.coding}
\end{align}
The affine constraint in \eqref{sparse.coding}, imposed on the
$\Set{\alpha_{ii'}}$ coefficients in representing $\mathbf{x}_{ii}$ via its
neighbors, is motivated by the affine nature of the tangent space
(Fig.~\ref{Fig:theta}). Moreover, the larger the distance of neighbor
$\mathbf{x}_{ii'}$ from $\mathbf{x}_{ii}$, the larger the weight
$\exp[{\norm{\mathbf{x}_{ii'} - \mathbf{x}_{ii}}/ \sigma_{\alpha}}]$, which in
turn penalizes severely the coefficient $\alpha_{ii'}$ by pushing it to values
close to zero. Step~\ref{Step:OnGrass:sample.covariance.matrix} computes the
sample covariance matrix
\begin{align}
  \hat{\vect{C}}_{x_i} \coloneqq \tfrac{1}{\lvert \mathscr{N}_{\text{NN}}(x_i)\rvert - 1}
  \sum\nolimits_{x_{i'}\in \mathscr{N}_{\text{NN}}(x_i)} (\vect{x}_{ii'} - \bar{\vect{x}}_i)
  (\vect{x}_{ii'} - \bar{\vect{x}}_i)^{\intercal}
  \,, \label{sample.covariance.matrix}
\end{align}
where
$\bar{\vect{x}}_i \coloneqq (1/\lvert \mathscr{N}_{\text{NN}}(x_i)\rvert)
\sum_{x_{i'}\in \mathscr{N}_{\text{NN}}(x_i)} \vect{x}_{ii'}$ denotes the sample
average of the neighbors of $\vect{x}_{ii}$. PCA is applied to
$\hat{\vect{C}}_{x_i}$ at step~\ref{Step:OnGrass:PCA} to compute the principal
eigenspace $\hat{S}_{x_i}$, which may be viewed as an approximation of the image
of the cluster (submanifold) $\mathcal{C}_k$, via the logarithm map, into the
tangent space $T_{x_i} \text{Gr}(\rho, mN)$ (see Fig.~\ref{Fig:theta}). Once
$\hat{S}_{x_i}$ is computed, the angle $\theta_{ii'}$ between vector
$\vect{x}_{ii'} - \vect{x}_{ii}$ and $\hat{\mathcal{S}}_{x_i}$ is also computed
at step~\ref{Step:OnGrass:theta} to extract angular information. The larger the
angle $\theta_{ii'}$ is, the less the likelihood for $x_{i'}$ to belong to
cluster $\mathcal{C}_k$. The information carried by both $\Set{\alpha_{ii'}}$
and the angles $\Set{\theta_{ii'}}$ is used to define the adjacency matrix
$\vect{W}$ at step~\ref{Step:OnGrass:adjacency.matrix}. The use of angular
information here, as well as in \cite{RMMM.arxiv, RMMM.AISTATS.15,
  slavakis2018clustering}, advances the boundary of state-of-the-art clustering
methods in the Grassmannian, where usually the weights of the adjacency matrix
are defined via the Grassmannian (geodesic) distance or sparse-coding
schemes~\cite{elhamifar2011sparse}.

\begin{algorithm}[!t]
  \DontPrintSemicolon
  \SetKwInOut{input}{Input}
  \SetKwInOut{parameters}{Parameters}
  \SetKwInOut{output}{Output}
  \SetKwBlock{initial}{Initialization}{}

  \input{Time series
    $\Set{(\prescript{}{\nu}{y}_{t})_{t\in \Integer}}_{\nu\in\mathcal{N}}$.}

  \output{Clusters $\Set{\mathcal{C}_k}_{k=1}^K$.}

  \BlankLine

  \SetKwFunction{FMain}{MainModule}

  \SetKwProg{Pn}{Function}{:}{}
  
  \Pn{\FMain{$\Set{(\prescript{}{\mathcal{V}}{\vect{y}}_t)_{t\in\mathfrak{T}}}_{\mathcal{V}\in
        \mathfrak{N}}$} \label{Step:module.start}}{ Apply
    Alg.~\ref{Algo:Map.O.to.Grassmannian} to obtain features $\Set{x_i}_{i\in
      \mathfrak{I}}$.

    Apply Alg.~\ref{Algo:OnGrass} to map $\Set{x_i}_{i\in \mathfrak{I}}$ to
    clusters $\Set{\mathcal{C}_k}_{k=1}^K$.
      
    \KwRet $\Set{\mathcal{C}_k}_{k=1}^K$. \label{Step:module.end}
  }

  \uIf{``state clustering''\label{Step:state.clustering:begin}}{
    
    Set $(\mathcal{V}, \mathfrak{N}) \coloneqq (\mathcal{N}, \Set{\mathcal{N}})$
    and form data $(\prescript{}{\mathcal{N}}{\vect{y}}_t)_{t\in\mathfrak{T}}$
    according to
    Sec.~\ref{Sec:state.clustering}. \label{Step:Framework:State:form.data}
    
    Call \FMain{$(\prescript{}{\mathcal{N}}{\vect{y}}_t)_{t\in\mathfrak{T}}$} to
    identify clusters/states. \label{Step:Framework:State:obtain.clusters}
    
  }
  \uElseIf{``community detection'' \label{Step:community.detection:begin}}{
    
    Set $(\mathcal{V}, \mathfrak{N}) \coloneqq (\mathcal{N}, \Set{\mathcal{N}})$
    and form data $(\prescript{}{\mathcal{N}}{\vect{y}}_t)_{t\in\mathfrak{T}}$
    according to
    Sec.~\ref{Sec:state.clustering}. \label{Step:Framework:Comm:form.data}

    Call \FMain{$(\prescript{}{\mathcal{N}}{\vect{y}}_t)_{t\in\mathfrak{T}}$} to
    identify states, \ie, cluster the time horizon $\mathfrak{T}$ into a
    partition $\Set{\mathfrak{T}_j}_{j=1}^J$ s.t.\ data
    $\Set{(\prescript{}{\nu}{y}_{t})_{t\in\mathfrak{T}_j}}_{\nu\in\mathcal{N}}$
    belong to the same state. \label{Step:Framework:Comm:obtain.clusters}

      \For{$j=1$ \KwTo $J$\label{Step:comm.detection:identify.comms.start}}
      {%

        Set $(\mathcal{V}, \mathfrak{N}) \coloneqq (\nu, \mathcal{N})$ and form
        data
        $\Set{(\prescript{}{\nu}{\vect{y}}_t)_{t\in\mathfrak{T}_j}}_{\nu\in
          \mathcal{N}}$ according to Sec.~\ref{Sec:comm.task.clustering}.

        Call
        \FMain{$\Set{(\prescript{}{\nu}{\vect{y}}_t)_{t\in\mathfrak{T}_j}}_{\nu\in
            \mathcal{N}}$} to identify communities in state
        $j$.\label{Step:community.detection:end}
        
      }
  }
  \ElseIf{``subnetwork-state-sequence
    clustering''\label{Step:task.clustering:begin}}{

    Call lines~\ref{Step:Framework:Comm:form.data} and
    \ref{Step:Framework:Comm:obtain.clusters} to identify states, \ie,
    a partition $\Set{\mathfrak{T}_j}_{j=1}^J$ of
    $\mathfrak{T}$.\label{Step:Framework:Task:find.states.first}

    \For{$j=1$ \KwTo $J$\label{Step:task.clustering:extract.features.begin}}
    {

      Set $(\mathcal{V}, \mathfrak{N}) \coloneqq (\nu, \mathcal{N})$ and form
      data
      $\Set{(\prescript{}{\nu}{\vect{y}}_t)_{t\in\mathfrak{T}_j}}_{\nu\in
        \mathcal{N}}$ according to Sec.~\ref{Sec:comm.task.clustering}.
      
      Apply Alg.~\ref{Algo:Map.O.to.Grassmannian} to
      $\Set{(\prescript{}{\nu}{\vect{y}}_t)_{t\in\mathfrak{T}_j}}_{\nu\in
        \mathcal{N}}$ to obtain the Grassmannian features
      $\Set{[\prescript{j}{\nu}{\hat{\vect{O}}}_t]}_{t\in
        \mathfrak{T}_j}$. \label{Step:task.clustering:extract.features.end}

    }

    Form features
    $\Set{x_i}_{i\in \mathfrak{I}} \coloneqq \cup_{j=1}^J \cup_{\nu\in
      \mathcal{N}} \cup_{t\in \mathfrak{T}_j}
    [\prescript{j}{\nu}{\hat{\vect{O}}}_t]$.\label{Step:task.clustering:gather.features}

    Apply Alg.~\ref{Algo:OnGrass} to $\Set{x_i}_{i\in \mathfrak{I}}$ to identify
    clusters/tasks.\label{Step:task.clustering:end}
    
  }

  \caption{Clustering framework}
  \label{Algo:the.clustering.framework}
\end{algorithm}

To summarize, the clustering framework is presented in pseudo-code form in
Alg.~\ref{Algo:the.clustering.framework}. More specifically, the main module of
the framework, which is frequently utilized and contains
Algs.~\ref{Algo:Map.O.to.Grassmannian} and \ref{Algo:OnGrass}, is presented at
steps \ref{Step:module.start}--\ref{Step:module.end}. While the
``state-clustering'' part
(steps~\ref{Step:state.clustering:begin}--\ref{Step:Framework:State:obtain.clusters})
is quite straightforward, the ``community detection''
(steps~\ref{Step:community.detection:begin}--\ref{Step:community.detection:end})
and ``subnetwork-state-sequence clustering''
(steps~\ref{Step:task.clustering:begin}--\ref{Step:task.clustering:end})
comprise several steps. More specifically, in ``community detection''
(steps~\ref{Step:community.detection:begin}--\ref{Step:community.detection:end}),
states are first identified via
steps~\ref{Step:Framework:Comm:form.data}--\ref{Step:Framework:Comm:obtain.clusters}
and then communities are identified in
steps~\ref{Step:comm.detection:identify.comms.start}--\ref{Step:community.detection:end}
within each state. In ``subnetwork-state-sequence clustering''
(steps~\ref{Step:task.clustering:begin}--\ref{Step:task.clustering:end}), states
are again identified first in step~\ref{Step:Framework:Task:find.states.first},
the Grassmannian features are extracted in
steps~\ref{Step:task.clustering:extract.features.begin}--\ref{Step:task.clustering:extract.features.end},
all features are gathered as $\Set{x_i}_{i\in \mathfrak{I}}$ in
step~\ref{Step:task.clustering:gather.features}, and finally
Alg.~\ref{Algo:OnGrass} is applied to $\Set{x_i}_{i\in \mathfrak{I}}$ to
identify clusters/tasks in step~\ref{Step:task.clustering:end}.

To achieve a high accuracy clustering result, it is necessary to cluster states
first, before applying community detection and subnetwork-state-sequence
clustering. Without knowing the starting and ending points of different states,
there will be time-series vectors $\prescript{}{\nu}{\vect{y}}_{t}$ in
Alg.~\ref{Algo:Map.O.to.Grassmannian} which capture data from two consecutive
states, since $\prescript{}{\nu}{\vect{y}}_{t}$ takes the form of
$[\prescript{}{\nu}{y}_t, \ldots, \prescript{}{\nu}{y}_{t+ \texttt{Buff}_{\nu}
  -1}]^{\intercal}$. Features corresponding to those vectors will decrease the
clustering accuracy since the extracted features do not correspond to any actual
state or task.

The main computational burden comes from the module of steps
\ref{Step:module.start}--\ref{Step:module.end} in
Alg.~\ref{Algo:the.clustering.framework}. If $\mathfrak{I}_{\mathcal{V}}$
denotes the points in the Grassmannian, the computational complexity for
computing features $\Set{x_i}_{i\in\mathfrak{I}_{\mathcal{V}}}$ in
Alg.~\ref{Algo:Map.O.to.Grassmannian} is
$\mathcal{O}(|\mathfrak{I}_{\mathcal{V}} |\mathcal{C}_{\Kprod})$, where
$\mathcal{C}_{\Kprod}$ denotes the cost of computing
$\bm{\mathcal{F}}_{t+1} \Kprod \bm{\mathcal{B}}_t^{\intercal}$, which includes
SVD computations. In Alg.~\ref{Algo:OnGrass}, the complexity for computing the
$\mathscr{N}_{\text{NN}}(x_i)$ nearest neighbors of $x_i$ is
$\mathcal{O}(|\mathfrak{I}_{\mathcal{V}}|\mathcal{C}_{\text{dist}} +
\mathscr{N}_{\text{NN}}\log|\mathfrak{I}_{\mathcal{V}}|)$, where
$\mathcal{C}_{\text{dist}}$ denotes the cost of computing the Riemannian
distance between any two points, and
$\mathscr{N}_{\text{NN}}\log|\mathfrak{I}_{\mathcal{V}}|$ refers to the cost of
finding the $\mathscr{N}_{\text{NN}}$ nearest neighbors of $x_i$.
Step~\ref{Step:OnGrass:sparse.coding} of Alg.~\ref{Algo:OnGrass} is a
sparsity-promoting optimization task of (6) and let $\mathcal{C}_{\text{SC}}$
denotes the complexity to solve it. Under
$\mathcal{M} \coloneqq \text{Gr}(\rho, mN)$, step~\ref{Step:OnGrass:PCA} of
Alg.~\ref{Algo:OnGrass} involves the computation of the eigenvectors of the
sample covariance matrix $\hat{\vect{C}}_{x_i}$, with complexity of
$\mathcal{O}(\dim\mathcal{M} + K_{\text{NN}}^3)$. In
step~\ref{Step:OnGrass:theta}, the complexity for computing empirical geodesic
angles is
$\mathcal{O}[|\mathfrak{I}_{\mathcal{V}}| (\mathcal{C}_{\log} +
\dim\mathcal{M})]$, where $\mathcal{C}_{\log}$ is the complexity of computing
the logarithm map $\log_{x_i}(\cdot)$~\cite{slavakis2018clustering}. For the
last step of Alg.~\ref{Algo:OnGrass}, the exact complexity of Louvain method is
not known but the method seems to run in time
$\mathcal{O}(|\mathfrak{I}_{\mathcal{V}}| \log|\mathfrak{I}_{\mathcal{V}}|)$
with most of the computational effort spent on modularity optimization at first
level, since modularity optimization is known to be NP-hard
\cite{de2011generalized}. To summarize, the complexity of
Alg.~\ref{Algo:OnGrass} is
$\mathcal{O} [|\mathfrak{I}_{\mathcal{V}}|^2(\mathcal{C}_{\text{dist}} +
\mathcal{C}_{\log} + \dim\mathcal{M}) +
(K_{\text{NN}}+1)|\mathfrak{I}_{\mathcal{V}}| \log|\mathfrak{I}_{\mathcal{V}}| +
|\mathfrak{I}_{\mathcal{V}}| (\dim\mathcal{M} + K_{\text{NN}}^3)]$.

\section{Numerical Tests}\label{Sec:Tests}

This section validates the proposed framework on synthetic and real data. First,
the competing clustering algorithms are briefly described.

\subsection{Competing Algorithms}

\subsubsection{Sparse Manifold Clustering and Embedding (SMCE)~\cite{elhamifar2011sparse}}

Each point on the Grassmannian is described by a sparse affine combination of
its neighbors. The computed sparse weights define the entries of a similarity
matrix, which is subsequently used to identify data-cluster associations. SMCE
does not utilize any angular information, as step~\ref{Step:OnGrass:theta} of
Alg.~\ref{Algo:OnGrass} does.

\subsubsection{Interaction K-means with PCA (IKM-PCA)~\cite{vijay2015brain}}

IKM is a clustering algorithm based on the classical K-means and Euclidean
distances within a properly chosen feature space. To promote time-efficient
solutions, the classical PCA is employed as a dimensionality-reduction tool for
feature-subset selection.

\subsubsection{Graph-shift-operator estimation (GOE)~\cite{Segarra:17}}

The graph shift operator is a symmetric matrix capturing the network's
structure, \ie topology. There are widely adopted choices of graph shift
operators, including the adjacency and Laplacian matrices, or their various
degree-normalized counterparts. An estimation algorithm in~\cite{Segarra:17}
computes the optimal graph shift operator via convex optimization. The computed
graph shift operator is fed to a spectral-clustering module to identify
communities within a single brain state, since \cite{Segarra:17} assumes
stationary time-series data.

\subsubsection{3D-Windowed Tensor Approach (3D-WTA)~\cite{al2017tensor}}

3D-WTA was originally introduced for community detection in dynamic networks by
applying tensor decompositions onto a sequence of adjacency matrices indexed
over the time axis. 3D-WTA was modified in~\cite{al2018tensor} to accommodate
multi-layer network structures. High-order SVD (HOSVD) and high-order orthogonal
iteration (HOOI) are used within a pre-defined sliding window to extract
subspace information from the adjacency matrices. The ``asymptotic-surprise''
metric is used as the criterion to determine the number of clusters. 3D-WTA is
capable of performing both state clustering and community detection.

SMCE, 3D-WTA and the classical K-means will be compared against
Alg.~\ref{Algo:the.clustering.framework} on state clustering. SMCE, IKM-PCA,
3D-WTA, GOA and K-means will be used in community detection. Since none of
IKM-PCA, GOA and 3D-WTA can perform subnetwork-state-sequence clustering across
multiple states, only the results of Alg.~\ref{Algo:the.clustering.framework}
and SMCE are reported. To ensure fair comparisons, the parameters of all methods
were carefully tuned to reach optimal performance for every scenario at hand.

In the following discussion, tags K-ARMA[S] and K-ARMA[M] denote the proposed
framework whenever a single and multiple kernel functions are employed,
respectively. In the case where the linear kernel is used, the K-ARMA method
boils down to the ARMA method of~\cite{slavakis2018clustering}.

The evaluation of all methods was based on the following three criteria: 1)
Clustering accuracy, defined as the number of correctly clustered data points
(ground-truth labels are known) over the total number of points; 2) normalized
mutual information (NMI)~\cite{schutze2008introduction}; and 3) the classical
confusion matrix~\cite{Fawcett.ROC}, with true-positive ratio (TPR),
false-positive ratio (FPR), true-negative ratio (TNR), and false-negative ratio
(FNR), in the case where the number of clusters to be identified is equal to
two. In what follows, every numerical value of the previous criteria is the
uniform average of $20$ independently performed tests for the particular
scenario at hand.

\subsection{Synthetic Data}

\subsubsection{fMRI Data}\label{Sec:synthetic.fMRI}

\begin{figure}[!t]
  \centering
  \subfloat[State 1]{\includegraphics[width =
    .25\linewidth]{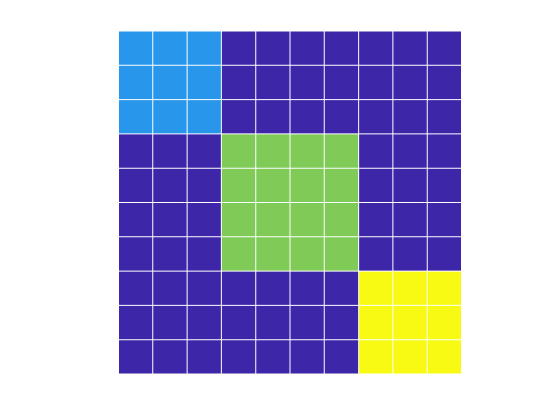}} 
  \subfloat[State 2]{\includegraphics[width =
    .25\linewidth]{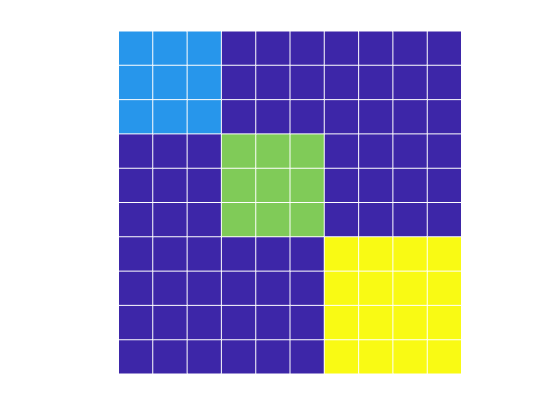}} 
  \subfloat[State 3]{\includegraphics[width =
    .25\linewidth]{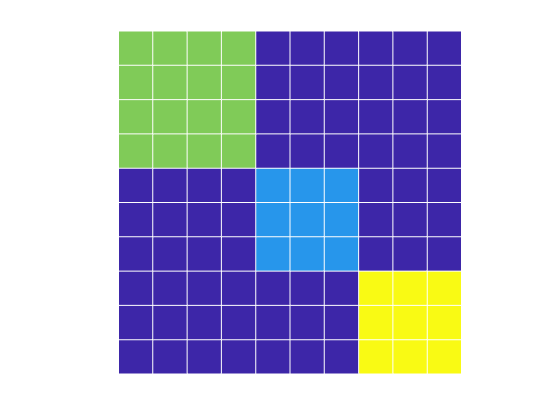}} 
  \subfloat[State 4]{\includegraphics[width =
    .25\linewidth]{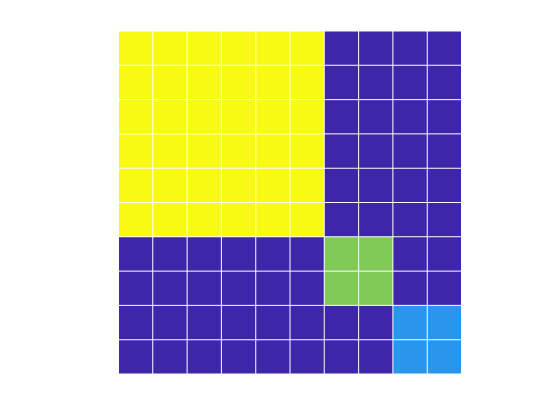}}\\ 
  \subfloat[BOLD time series of node \#2, dataset \#5]{\includegraphics[width =
    1\linewidth]{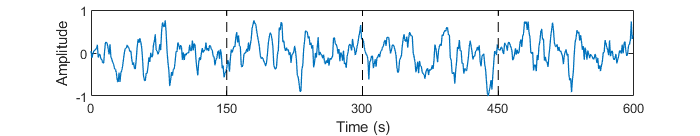}\label{Fig:fMRI.BOLD}}
  \caption{Synthetic data generated by the Matlab SimTB
    toolbox~\cite{allen2014tracking}. (a)-(d) Noiseless and outlier-free
    connectivity matrices corresponding to four network states. Nodes that share
    the same color cooperate to perform a common subnetwork
    task.}\label{Fig:synthetic.fMRI}
\end{figure}

Data were generated by the open-source Matlab SimTB
toolbox~\cite{allen2014tracking}. A $10$-node network is considered that
transitions successively between $4$ distinct network states. Each state
corresponds to a certain connectivity matrix, generated via the following
path. Each connectivity matrix, fed to the SimTB toolbox, is modeled as the
superposition of three matrices: 1) The ground-truth (noiseless) connectivity
matrix (\cf Fig.~\ref{Fig:synthetic.fMRI}), where nodes sharing the same color
belong to the same cluster and collaborate to perform a common subnetwork task;
2) a symmetric matrix whose entries are drawn independently from a zero-mean
Gaussian distribution with standard deviation $\sigma$ to model noise; and 3) a
symmetric outlier matrix where $36$ entries are equal to $\mu$ to account for
outlier neural activity.

Different states may share different outlier matrices, controlled by
$\mu$. Aiming at extensive numerical tests, six datasets were generated
(corresponding to the columns of
Table~\ref{Table:synthetic.fMRI:state.clustering}) by choosing six pairs of
parameters $(\mu, \sigma)$ in the modeling of the connectivity matrices and the
SimTB toolbox. Datasets 1, 2 and 3 (D1, D2 and D3) were created without
outliers, while datasets 4, 5 and 6 (D4, D5 and D6) include outlier matrices
with different $\mu$s in different
states. Table~\ref{Table:parameters.of.synthetic.fMRI:state.clustering} details
the parameters of those six datasets. Driven by the previous connectivity
matrices, the SimTB toolbox generates BOLD time
series~\cite{ogawa1990brain}. Each state contributes $150$ signal samples, for a
total of $4\times 150 = 600$ samples, to every nodal time series, \eg
Fig.~\ref{Fig:fMRI.BOLD}.

\begin{figure}[!t]
  \centering
  \subfloat[State 1]{\includegraphics[width=.25\linewidth]{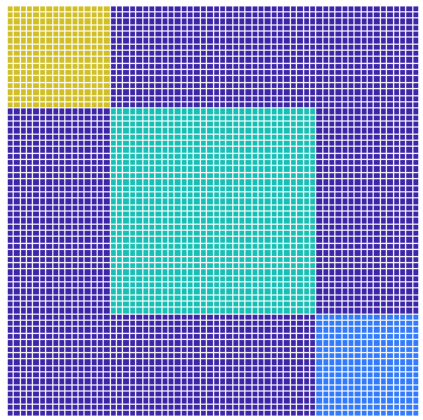}\label{Fig:EEG.state1}}
  \qquad
  \subfloat[State 2]{\includegraphics[width=.25\linewidth]{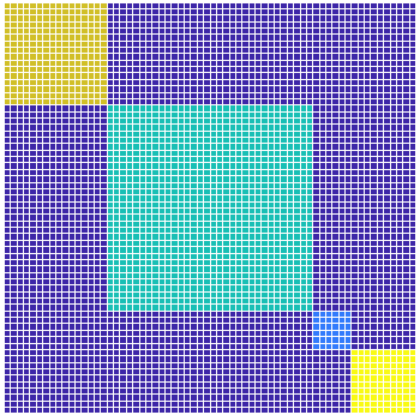}\label{Fig:EEG.state2}}\\
  \subfloat[D1S1]{\includegraphics[width=.25\linewidth]{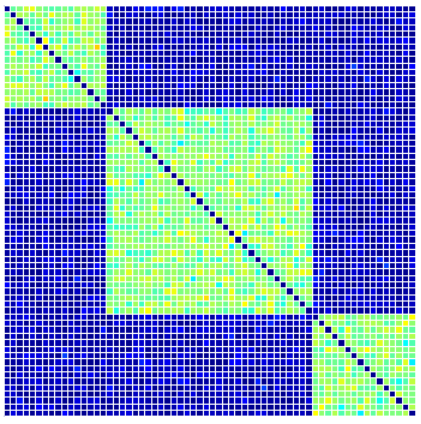}\label{Fig:EEG.conn.matrix.D1S1}}
  \qquad
  \subfloat[D1S2]{\includegraphics[width=.25\linewidth]{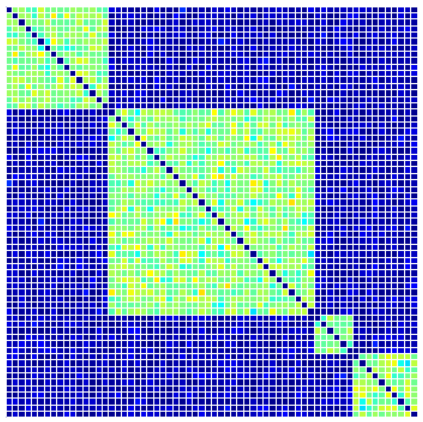}}\\
  \subfloat[D2S1]{\includegraphics[width=.25\linewidth]{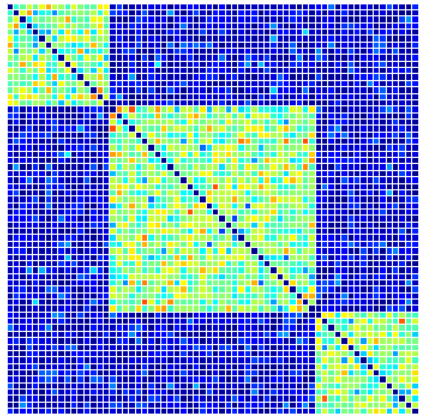}}
  \qquad
  \subfloat[D2S2]{\includegraphics[width=.25\linewidth]{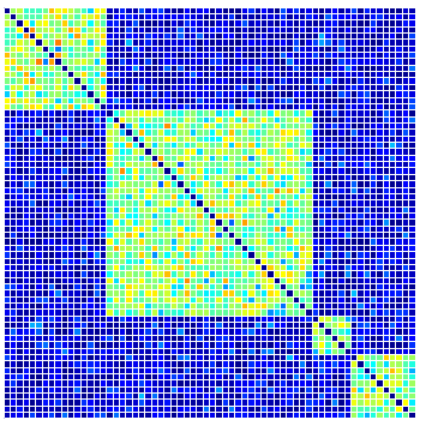}}\\
  \subfloat[D3S1]{\includegraphics[width=.25\linewidth]{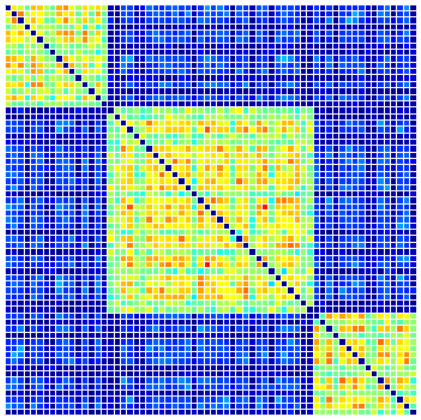}}
  \qquad
  \subfloat[D3S2]{\includegraphics[width=.25\linewidth]{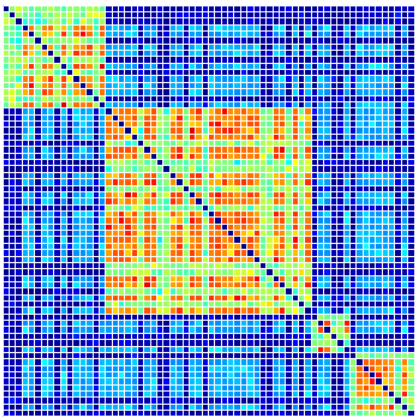}\label{Fig:EEG.conn.matrix.D3S2}}\\
  \subfloat[EEG time series of node \#27, dataset \#2]{\includegraphics[width =
    1\linewidth]{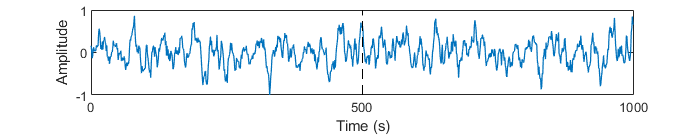}\label{Fig:EEG.signal}}
  \caption{Synthetic data generated by the Virtual Brain
    toolbox~\cite{sanz2013virtual}. (a) and (b) Task connectivity matrices
    without noise for each state. There are three communities/tasks for state~1 and
    four tasks in state~2. (c)-(h) are connectivity matrices for each state of three
    datasets. (g) is the time series of 27th node of
    dataset~2.}\label{Fig:synthetic.EEG}
\end{figure}

\begin{table*}[tp]
  \centering
  \resizebox{.9\textwidth}{!}{%
  \begin{threeparttable}
    \caption{Synthetic fMRI Data: State clustering}
    \label{Table:synthetic.fMRI:state.clustering}
    \begin{tabular}{c|c|c|c||c|c|c||c|c|c||c|c|c}
      \toprule
      \multirow{3}{*}{Methods}
      & \multicolumn{6}{c}{Without Outliers}
      & \multicolumn{6}{c}{With Outliers}\cr
        \cmidrule(lr){2-7} \cmidrule(lr){8-13}
      & \multicolumn{3}{c}{Clustering Accuracy}
      & \multicolumn{3}{c}{NMI}
      & \multicolumn{3}{c}{Clustering Accuracy}
      & \multicolumn{3}{c}{NMI} \cr \cmidrule(lr){2-4} \cmidrule(lr){5-7} \cmidrule(lr){8-10}
        \cmidrule(lr){11-13} %
         \cmidrule(lr){3-5} \cmidrule(lr){6-8} \cmidrule(lr){9-11} \cmidrule(lr){11-13}
& D1 & D2 & D3 & D1 & D2 & D3 & D4 & D5 & D6 & D4 & D5 & D6 \cr\midrule
ARMA&0.969&0.805&0.640&0.948&0.766&0.596&0.944&0.743&0.589&0.860&0.627&0.340\cr
K-ARMA[S]&\textbf{1}&0.824&0.671&\textbf{1}&0.791&0.622&0.983&0.775&0.599&0.930&0.651&0.379\cr
K-ARMA[M]&\textbf{1}&\textbf{0.839}&\textbf{0.708}&\textbf{1}&\textbf{0.808}&\textbf{0.641}&\textbf{0.992}&\textbf{0.800}&\textbf{0.626}&\textbf{0.967}&\textbf{0.689}&\textbf{0.435}\cr
3DWTA&\textbf{1}&0.792&0.603&\textbf{1}&0.735&0.556&0.943&0.721&0.517&0.872&0.558&0.281\cr              
SMCE&0.920&0.784&0.583&0.887&0.673&0.480&0.883&0.712&0.508&0.713&0.562&0.246\cr
Kmeans&0.866&0.670&0.402&0.800&0.590&0.307&0.768&0.621&0.337&0.476&0.403&0.168\cr
                                                                          \bottomrule
    \end{tabular}
  \end{threeparttable}
  }
\end{table*}

\begin{figure}[htpb]
\centering
  \subfloat[Datasets 1, 2, 3]{
  \includegraphics[width = .7\columnwidth]{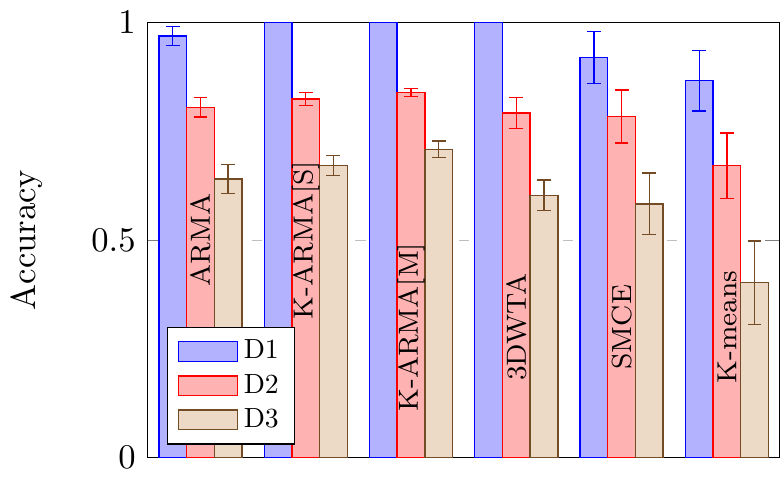}
     
  }

  \subfloat[Datasets 4, 5, 6]{
  \includegraphics[width = .7\columnwidth]{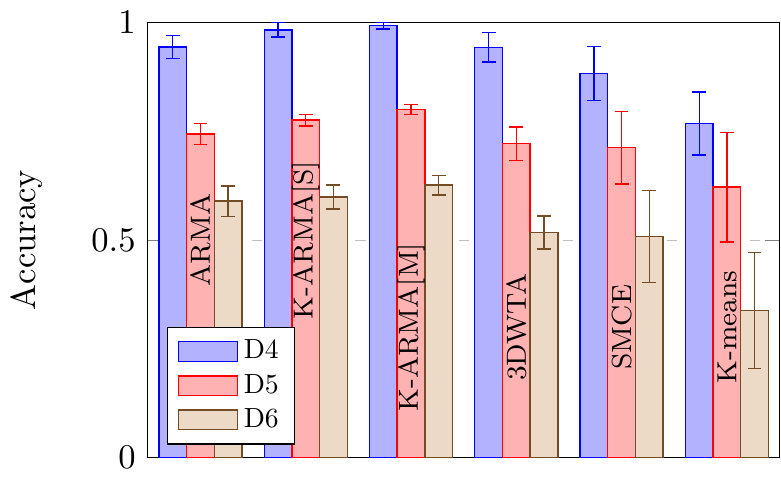}
  }
  \caption{State-clustering results of synthetic fMRI datasets. (a) Data without
    an independent event; (b) Data with an independent
    event.} \label{Fig:synthetic.fMRI:state.clustering} 
\end{figure}

Table~\ref{Table:synthetic.fMRI:state.clustering} demonstrates the results of
state clustering. The parameters used for ARMA and K-ARMA are: $N \coloneqq 30$,
$m \coloneqq 2$, $\rho \coloneqq 2$, $\tau_{f} \coloneqq 60$,
$\tau_b \coloneqq 20$. The Gaussian kernel
$\kappa_{\text{G}; 0.8}(\cdot, \cdot)$ (\cf App.~\ref{App:RKHS}) is used in the
single-kernel method K-ARMA[S], while kernel
$\kappa(\cdot, \cdot) \coloneqq 0.6\,\kappa_{\text{G}; 0.8}(\cdot, \cdot) +
0.4\,\kappa_{\text{L}; 1}(\cdot, \cdot)$ is used in the K-ARMA[M] case since it
performed the best among other choices of kernel
functions. Fig.~\ref{Fig:synthetic.fMRI:state.clustering} depicts also the
standard deviations of the results of
Table~\ref{Table:synthetic.fMRI:state.clustering}, computed after performing
independent repetitions of the same test. Standard deviation of all algorithms
increase when the strength of the noisy matrix increases. For dataset D1,
K-ARMA[S], K-ARMA[M] and 3DWTA reach $100\%$ accuracy; for other datasets,
K-ARMA[M] exhibits the highest accuracy and the smallest standard deviation.

3DWTA, K-ARMA[S] and K-ARMA[M] achieved perfect score ($100\%$) for both the
clustering-accuracy and NMI metrics on dataset $1$. Among all methods, K-ARMA[M]
scores the highest clustering accuracy and NMI over all six datasets. It can be
observed by Table~\ref{Table:synthetic.fMRI:state.clustering} that the existence
of outliers affects negatively the ability of all methods to cluster data. The
main reason is that the algorithms tend to detect outliers and gather those in
clusters different from the nominal ones. Ways to reject those outliers are
outside of the scope of this study and will be provided in a future
publication. 

\begin{table*}[tp]
  \centering
   \resizebox{.9\textwidth}{!}{%
  \begin{threeparttable}
    \caption{Synthetic fMRI data: Community detection}
    \label{Table:synthetic.fMRI:community.detection}
    \begin{tabular}{c|c|c|c||c|c|c||c|c|c||c|c|c}
      \toprule
      \multirow{3}{*}{Methods}
      & \multicolumn{6}{c}{Without Outliers}
      & \multicolumn{6}{c}{With Outliers}\cr
        \cmidrule(lr){2-7} \cmidrule(lr){8-13}
      & \multicolumn{3}{c}{\begin{tabular}{@{}c@{}}Clustering\\ Accuracy\end{tabular}}
      & \multicolumn{3}{c}{NMI}
      & \multicolumn{3}{c}{\begin{tabular}{@{}c@{}}Clustering\\ Accuracy\end{tabular}}
      & \multicolumn{3}{c}{NMI}\cr
        \cmidrule(lr){2-4} \cmidrule(lr){5-7} \cmidrule(lr){8-10} \cmidrule(lr){11-13}
      & D1    & D2    &D3   & D1    & D2    &D3   & D4    & D5   &D6    & D4    & D5    &D6\cr
                                                                \midrule
    ARMA      &\textbf{1} &0.960 &0.842 &\textbf{1} &0.876 &0.775 &0.973 &0.910 &0.817 &0.940 &0.793 &0.664\cr
    K-ARMA[S] &\textbf{1} &\textbf{1} &0.915 &\textbf{1} &\textbf{1} &0.838 &\textbf{1} &0.942 &0.852 &\textbf{1} &0.864 &0.710\cr
    K-ARMA[M] &\textbf{1} &\textbf{1} &\textbf{0.945} &\textbf{1} &\textbf{1} &\textbf{0.907} &\textbf{1} &\textbf{0.958} &\textbf{0.879}
      &\textbf{1} & \textbf{0.892} &\textbf{0.803}\cr
    3DWTA     &\textbf{1} &0.941 &0.839 &\textbf{1} &0.927 &0.754 &0.925 &0.863 &0.799 &0.842 &0.780 &0.638 \cr  
    SMCE      &0.975 &0.929 &0.827 &0.902 &0.865 &0.691 &0.909 &0.773 &0.745 &0.769 &0.647 &0.563 \cr
    GOE       &\textbf{1} &0.933 &0.809 &\textbf{1} &0.915 &0.655 &0.918 &0.740
    &0.684 &0.833 &0.652 &0.409 \cr
    IKM-PCA   &0.948 &0.907 &0.791 &0.890 &0.814 &0.629 &0.892 &0.756 &0.712 &0.738 &0.551 &0.486\cr
    Kmeans    &0.908 &0.876 &0.725 &0.810 &0.729 &0.547 &0.843 &0.672 &0.605 &0.620 &0.391 &0.314\cr
    \bottomrule
    \end{tabular}
    
  \end{threeparttable}
}
\end{table*}

\begin{figure}[htpb]
  \centering
  \subfloat[Datasets 1, 2, 3]{
    \includegraphics[width = .7\columnwidth]{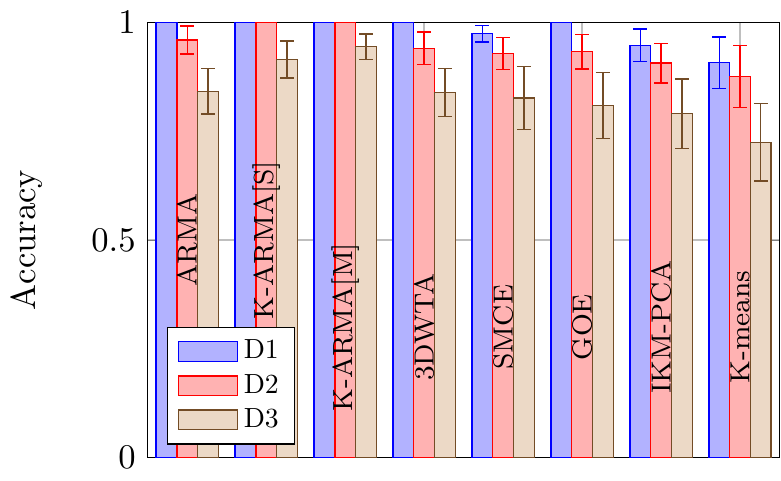}
  }
  
\subfloat[Datasets 4, 5, 6]{
 \includegraphics[width = .7\columnwidth]{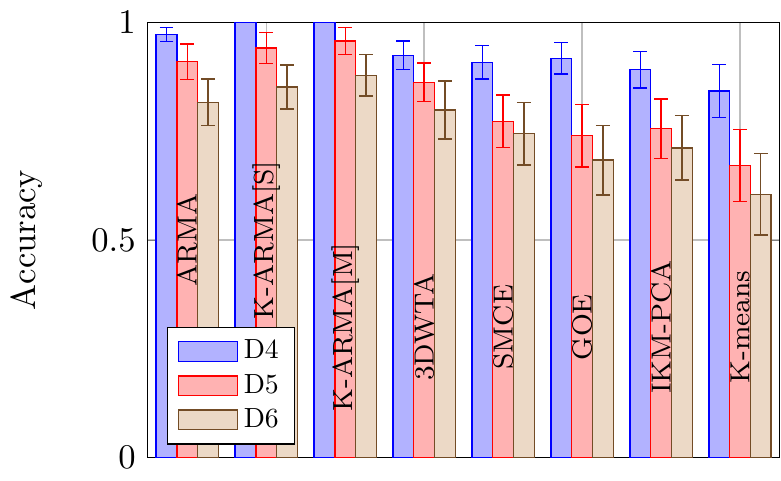}
     }
     \caption{Community detection results of synthetic fMRI datasets. (a) Data
       without an independent event; (b) Data with an independent
       event.} \label{Fig:synthetic.fMRI:community detection}
\end{figure}

Table~\ref{Table:synthetic.fMRI:community.detection} presents the results of
community detection. The numerical values in
Table~\ref{Table:synthetic.fMRI:community.detection} stand for the average
values over the $4$ states for each one of the datasets. Parameters of ARMA and
K-ARMA were set as follows: $N \coloneqq 30$,
$\texttt{Buff}_{\nu} \coloneqq 20$, $m \coloneqq 3$, $\rho \coloneqq 2$,
$\tau_f \coloneqq 50$, $\tau_b \coloneqq 10$. In K-ARMA[S], the utilized kernel
function is $\kappa_{\text{G}; 0.5}(\cdot, \cdot)$, while in K-ARMA[M] the
kernel is defined as
$\kappa(\cdot, \cdot) \coloneqq 0.5\, \kappa_{\text{G}; 0.5}(\cdot, \cdot) +
0.5\, \kappa_{\text{L}0; 1}(\cdot, \cdot)$ (\cf
App.~\ref{App:RKHS}). Table~\ref{Table:synthetic.fMRI:community.detection}
demonstrates that K-ARMA[M] consistently outperforms all other methods across
all datasets and even for the case where outliers contaminate the
data. Fig.~\ref{Fig:synthetic.fMRI:community detection} depicts also the
standard deviations of the results of
Table~\ref{Table:synthetic.fMRI:community.detection}. ARMA, K-ARMA[S], K-ARMA[M]
and 3DWTA score $100\%$ accuracy for dataset D1, while K-ARMA[S] and K-ARMA[M]
show $100\%$ accuracy for dataset D4. K-ARMA[M] shows the highest accuracy on
all other datasets.

\begin{table*}[tp]
  \centering
  \resizebox{.9\textwidth}{!}{%
  \begin{threeparttable}
    \caption{Synthetic fMRI data: Subnetwork state sequences}
    \label{Table:synthetic.fMRI:task.clustering}
 \begin{tabular}{c|c|c|c||c|c|c||c|c|c||c|c|c}
      \toprule
      \multirow{3}{*}{Methods}
      & \multicolumn{6}{c}{Without Outliers}
      & \multicolumn{6}{c}{With Outliers}\cr
        \cmidrule(lr){2-7} \cmidrule(lr){8-13}
      & \multicolumn{3}{c}{\begin{tabular}{@{}c@{}}Clustering\\ Accuracy\end{tabular}}
      & \multicolumn{3}{c}{NMI}
      & \multicolumn{3}{c}{\begin{tabular}{@{}c@{}}Clustering\\ Accuracy\end{tabular}}
      & \multicolumn{3}{c}{NMI}\cr
        \cmidrule(lr){2-4} \cmidrule(lr){5-7} \cmidrule(lr){8-10} \cmidrule(lr){11-13}
      & D1    & D2    &D3   & D1    & D2    &D3   & D4    & D5   &D6    & D4    & D5    &D6\cr
                                                                \midrule
    ARMA      &\textbf{1} &0.816 &0.749 &\textbf{1} &0.767 &0.684 &0.928 &0.701 &0.633 &0.874 &0.484 &0.355\cr
    K-ARMA[S] &\textbf{1} &0.856 &0.781 &\textbf{1} &0.791 &0.702 &0.956 &0.728 &0.664 &0.913 &0.534 &0.410\cr
    K-ARMA[M] &\textbf{1} &\textbf{0.884} &\textbf{0.817} &\textbf{1} &\textbf{0.821} &\textbf{0.739} &\textbf{1} &\textbf{0.757} &\textbf{0.721}
      &\textbf{1} & \textbf{0.602} &\textbf{0.485}\cr
    SMCE      &0.936 &0.792 &0.691 &0.804 &0.617 &0.495 &0.851 &0.665 &0.580 &0.785 &0.416 &0.318 \cr
    \bottomrule
    \end{tabular}
    
  \end{threeparttable}
}
\end{table*}

\begin{figure}
  \centering
  \subfloat[Datasets 1, 2, 3]{
   \includegraphics[width = .7\columnwidth]{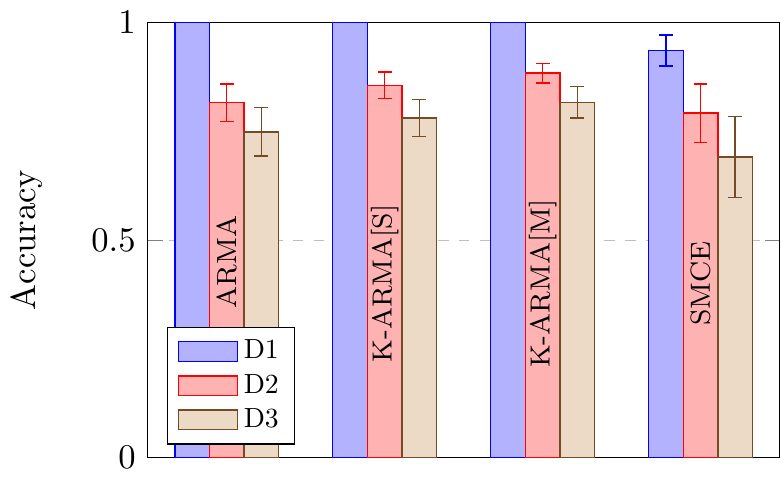}
  }
  
\subfloat[Datasets 4, 5, 6]{
\includegraphics[width = .7\columnwidth]{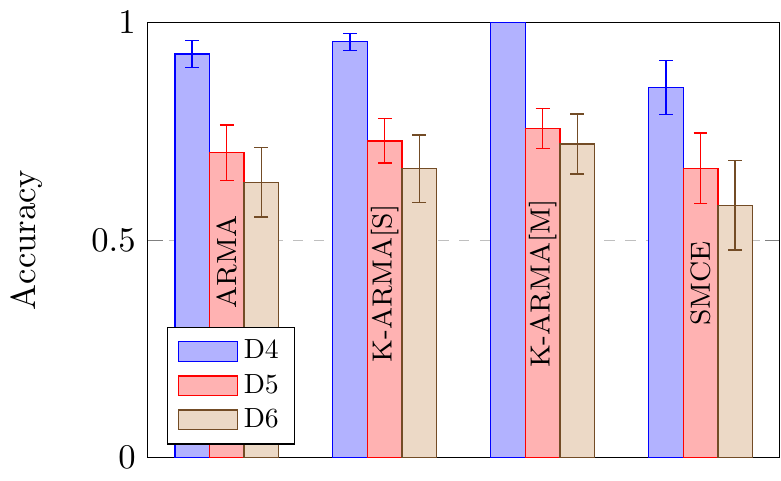}
    }
    \caption{Subnetwork-state-sequence clustering results of synthetic fMRI
      datasets. (a) Data without an independent event; (b) Data with an
      independent event.} \label{Fig:synthetic.fMRI:task tracking}
\end{figure}

Table~\ref{Table:synthetic.fMRI:task.clustering} illustrates the results of task
clustering on $4$ synthetic-fMRI datasets: D1, D2, D3, D4, D5 and D6. The
parameters of ARMA and Kernel ARMA were set as follows: $N \coloneqq 20$,
$\texttt{Buff}_{\nu} \coloneqq 50$, $m \coloneqq 3$, $\rho \coloneqq 3$,
$\tau_f \coloneqq 45$, $\tau_b \coloneqq 5$. The kernel functions used in
K-ARMA[S] and K-ARMA[M] are identical to those employed in
Table~\ref{Table:synthetic.fMRI:community.detection}. Similarly to the previous
cases, K-ARMA[M] outperforms all other methods across all datasets and scenarios
on both clustering accuracy and NMI. Fig.~\ref{Fig:synthetic.fMRI:task tracking}
depicts also the standard deviations of the results of
Table~\ref{Table:synthetic.fMRI:task.clustering}. ARMA, K-ARMA[S] and K-ARMA[M]
score $100\%$ accuracy on dataset D1. K-ARMA[M] shows the highest accuracy with
the smallest standard deviation on all other datasets.

Table~\ref{Table:parameters.of.synthetic.fMRI:state.clustering} provides the
parameters $\mu$ and $\sigma$ used to generate noise matrices and symmetric
matrices to simulate outlier neural activities. By choosing different
combinations of $(\mu, \sigma)$, $6$ different synthetic fMRI datasets were
created.

\begin{table*}
  \caption{Parameters $(\mu,\sigma)$ used to generate synthetic BOLD time
    series}
  \label{Table:parameters.of.synthetic.fMRI:state.clustering}
  \centering
  \begin{tabular}{|c|c|c|c|c|}
    \hline
    Dataset  & State 1 & State 2 & State 3 & State 4\\
    \hline
    1    &$(0,-10\text{dB})$ &$(0,-10\text{dB})$ &$(0,-10\text{dB})$
    &$(0,-10\text{dB})$ \\ 
    \hline
    2    &$(0,-8\text{dB})$ &$(0,-8\text{dB})$ &$(0,-8\text{dB})$
    &$(0,-8\text{dB})$ \\ 
    \hline
    3    &$(0,-6\text{dB})$ &$(0,-6\text{dB})$ &$(0,-6\text{dB})$
    &$(0,-6\text{dB})$ \\ 
    \hline
    4    &$(0.2,-10\text{dB})$ &$(0.3,-10\text{dB})$ &$(0.4,-10\text{dB})$
    &$(0.5,-10\text{dB})$ \\ 
    \hline
    5    &$(0.2,-8\text{dB})$ &$(0.3,-8\text{dB})$ &$(0.4,-8\text{dB})$
    &$(0.5,-8\text{dB})$ \\  
    \hline
    6    &$(0.2,-6\text{dB})$ &$(0.3,-6\text{dB})$ &$(0.4,-6\text{dB})$
    &$(0.5,-6\text{dB})$ \\ 
    \hline
  \end{tabular}
\end{table*}
 
\subsubsection{EEG Data}\label{Sec:synthetic.EEG}

Synthetic EEG data were generated by the open-source Virtual Brain (VB)
toolbox~\cite{sanz2013virtual}. A $60$-node network is considered that
transitions between two states, with noiseless and outlier-free connectivity
matrices depicted in Figs.~\ref{Fig:EEG.state1} and \ref{Fig:EEG.state2}. It is
worth noticing that the number of communities in Fig.~\ref{Fig:EEG.state1} is
$3$, while $4$ in Fig.~\ref{Fig:EEG.state2}. Similarly to the previous fMRI
case, every connectivity matrix, which is fed to the VB
toolbox~\cite{sanz2013virtual}, is the superposition of three matrices: The
ground-truth matrix (\cf Figs.~\ref{Fig:EEG.state1} and \ref{Fig:EEG.state2}),
the ``noise'' and the ``outlier'' matrices. Three scenarios/datasets are
considered, with the noisy and outlier-contaminated connectivity matrices
illustrated in
Figs.~\ref{Fig:EEG.conn.matrix.D1S1}--\ref{Fig:EEG.conn.matrix.D3S2}. Each state
contributes $500$ signal samples, for a total of $2\times 500 = 1,000$ samples,
to every nodal time series, \eg Fig.~\ref{Fig:EEG.signal}.

\begin{table*}[tp]
  \centering
  \resizebox{\columnwidth}{!}{%
  \begin{threeparttable}
    \caption{Synthetic EEG data: State clustering}
    \label{Table:synthetic.EEG:state.clustering}
    \begin{tabular}{c|c|c|c||c|c|c}
    \toprule
    \multirow{3}{*}{Methods}&
    \multicolumn{3}{c}{Clustering Accuracy} & \multicolumn{3}{c}{NMI}\cr
    \cmidrule(lr){2-4} \cmidrule(lr){5-7}
              & D1 & D2 & D3 & D1 & D2 & D3   \cr
    \midrule
    ARMA   &\textbf{1} &\textbf{1} &0.944 &\textbf{1} &\textbf{1} &0.906 \cr
    K-ARMA[S] &\textbf{1} &\textbf{1} &0.963 &\textbf{1} &\textbf{1} &0.926 \cr
    K-ARMA[M] &\textbf{1} &\textbf{1} &\textbf{0.992} &\textbf{1} &\textbf{1}
      &\textbf{0.984} \cr 
    3DWTA &\textbf{1} &0.976 &0.939 &\textbf{1} &0.942 &0.896 \cr
    SMCE &0.952 &0.901 &0.804 &0.914 &0.852 &0.766 \cr
    Kmeans &0.879 &0.793 &0.732 &0.796 &0.761 &0.694 \cr
    \bottomrule
    \end{tabular}
  \end{threeparttable}
}
\end{table*}

\begin{figure}
  \centering
  \includegraphics[width = .7\columnwidth]{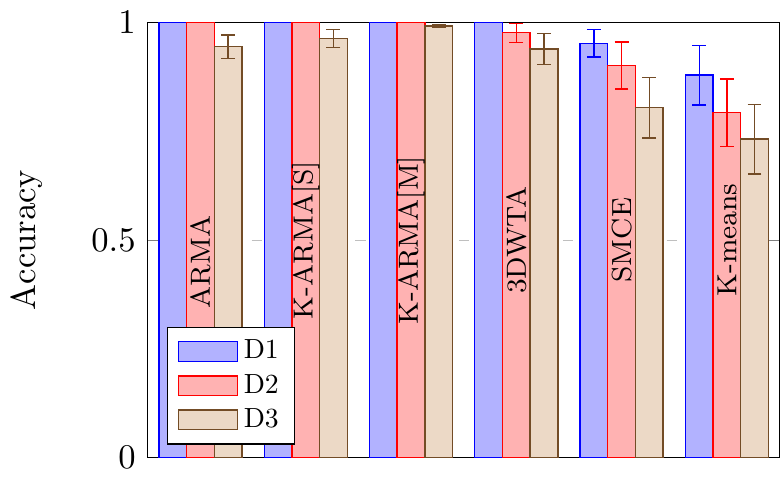}
  \caption{State-clustering results of synthetic EEG data.}
  \label{Fig:synthetic.EEG:state.clustering}
\end{figure}

The results of state clustering are shown in
Table~\ref{Table:synthetic.EEG:state.clustering}; standard deviations are also
included in Fig.~\ref{Fig:synthetic.EEG:state.clustering}. Parameters of ARMA
and K-ARMA were set as follows: $N \coloneqq 100$, $m \coloneqq
2$, $\rho \coloneqq 2$, $\tau_f \coloneqq 150$, $\tau_b \coloneqq 30$. The
reproducing kernel used are $\kappa_{\text{G}; 0.5}(\cdot, \cdot)$ for
K-ARMA[S], and
$\kappa(\cdot, \cdot) \coloneqq 0.5\, \kappa_{\text{G}; 0.5}(\cdot, \cdot) +
0.5\, \kappa_{\text{L}; 1}(\cdot, \cdot)$ for K-ARMA[M]. Furthermore, since
clustering on dataset 3 is performed between two states, the entries of the
confusion matrix for 3DWTA are $\text{TPR} = 0.911$, $\text{FPR} = 0.032$,
$\text{FNR} = 0.088$ and $\text{TNR} = 0.967$, while those for K-ARMA[S] are
$\text{TPR} = 0.955$, $\text{FPR} = 0.028$, $\text{FNR} = 0.044$ and
$\text{TNR} = 0.971$. 
Fig.~\ref{Fig:synthetic.EEG:state.clustering} depicts the standard deviations of
the results of Table~\ref{Table:synthetic.EEG:state.clustering}. Due to noise
and outliers, all algorithms show the highest accuracy on dataset 1 and the
lowest accuracy on dataset~3. ARMA,K-ARMA[S], K-ARMA[M] and 3DWTA score $100\%$
accuracy on dataset~1. ARMA, K-ARMA[S], and K-ARMA[M] exhibit $100\%$ accuracy
on dataset 2, while K-ARMA[M] scores the highest accuracy and the smallest
standard deviation on all other datasets.

\begin{table*}[tp]
  \centering
  \resizebox{\columnwidth}{!}{%
  \begin{threeparttable}
    \caption{Synthetic EEG data: Community detection}
    \label{Table:synthetic.EEG:comm.detection}
    \begin{tabular}{c|c|c|c|c|c|c}
    \toprule
      \multirow{3}{*}{Methods}
      & \multicolumn{3}{c}{Clustering accuracy}
      & \multicolumn{3}{c}{NMI}\cr
    \cmidrule(lr){2-4} \cmidrule(lr){5-7}
      & D1 & D2 & D3 & D1 & D2 & D3 \cr \midrule
    ARMA & \textbf{1} & 0.975 & 0.923 & \textbf{1} & 0.924 & 0.861 \cr
    K-ARMA[S] & \textbf{1} & \textbf{1} & 0.958 & \textbf{1} & \textbf{1} & 0.895 \cr
    K-ARMA[M] & \textbf{1} & \textbf{1} & \textbf{0.972} & \textbf{1} & \textbf{1} & \textbf{0.921} \cr
    3DWTA & \textbf{1} & \textbf{1} & 0.867 & \textbf{1} & \textbf{1} & 0.753 \cr 
    SMCE & \textbf{1}  &0.951 &0.833 & \textbf{1} &0.887 &0.706 \cr
    GOE  & \textbf{1}  &0.946 &0.785 & \textbf{1} &0.869 &0.684 \cr
    IKM-PCA & 0.961 & 0.913 & 0.811 & 0.935 & 0.847 & 0.652 \cr
    Kmeans & 0.894 & 0.808 & 0.696 & 0.836 & 0.713 & 0.530 \cr
    \bottomrule
    \end{tabular}
  \end{threeparttable}
  }
\end{table*}

\begin{figure}[htpb]
  \centering
  \includegraphics[width = .7\columnwidth]{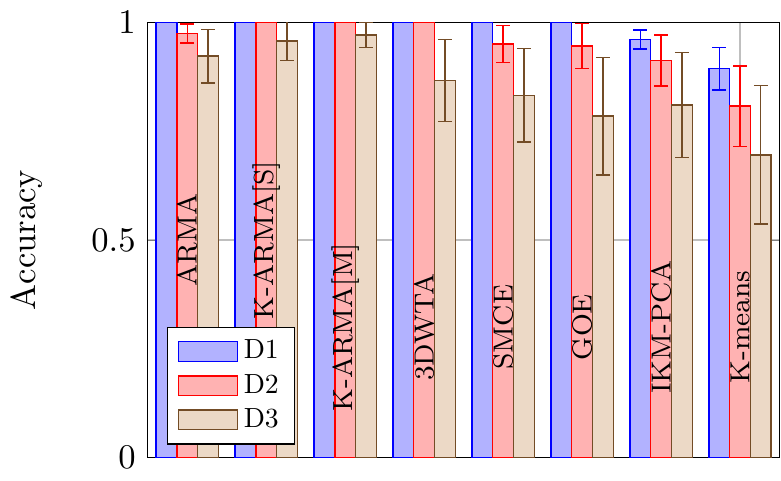}

  \caption{Community-detection results of synthetic EEG datasets.} 
  \label{Fig:synthetic.EEG:community detection}
\end{figure}

Table~\ref{Table:synthetic.EEG:comm.detection} illustrates the results of
community detection of synthetic EEG datasets, which include three time series
(D1--D3) across two states. The illustrated values are the average results over
the two network states per time series. The parameters of ARMA, K-ARMA[S] and
K-ARMA[M] were set as follows: $N \coloneqq 50$, $\texttt{Buff}_{\nu} = 50$,
$m \coloneqq 3$, $\rho \coloneqq 3$, $\tau_f \coloneqq 200$,
$\tau_b \coloneqq 10$. Moreover, $\kappa_{\text{G}; 0.5}(\cdot, \cdot)$ was used
as the reproducing kernel function in K-ARMA[S], while
$\kappa(\vect{x}, \vect{x}') \coloneqq 0.5\, \kappa_{\text{G}; 0.5}(\vect{x},
\vect{x}') + 0.5\, \kappa_{\text{L}; 0.8}(\vect{x}, \vect{x}')$ in
K-ARMA[M]. Fig.~\ref{Fig:synthetic.EEG:community detection} depicts also the
standard deviations of the results of
Table~\ref{Table:synthetic.EEG:comm.detection}. Due to noise and outliers, all
algorithms show their highest accuracies on dataset 1 and their lowest one on
dataset 3. ARMA, K-ARMA[S], K-ARMA[M] and 3DWTA show $100\%$ accuracy on dataset
1. K-ARMA[S], K-ARMA[M] and 3DWTA score $100\%$ accuracy on dataset 2. K-ARMA[M]
still shows the highest accuracy with the smallest standard deviation on all
other datasets.

\begin{table*}[t]
  \centering
  \resizebox{.9\textwidth}{!}{%
  \begin{threeparttable}
    \caption{Synthetic EEG Data: Subnetwork state sequences}
    \label{Table:synthetic.EEG:task.tracking}
    \begin{tabular}{c|c|c|c||c|c|c||c|c|c||c|c|c}
    \toprule
      \multirow{3}{*}{Methods}
      & \multicolumn{6}{c}{Unknown state label}
      & \multicolumn{6}{c}{Known state label}\cr
        \cmidrule(lr){2-7} \cmidrule(lr){8-13}
      & \multicolumn{3}{c}{Clustering accuracy}
      & \multicolumn{3}{c}{NMI}
      & \multicolumn{3}{c}{Clustering accuracy}
      & \multicolumn{3}{c}{NMI}\cr
    \cmidrule(lr){2-4} \cmidrule(lr){5-7} \cmidrule(lr){8-10} \cmidrule(lr){11-13}
      & D1 & D2 & D3 & D1 & D2 & D3 & D1 & D2 & D3 & D1 & D2 & D3\cr \midrule
    ARMA & \textbf{1} & 0.932 & 0.798 & \textbf{1} & 0.882 & 0.710 & \textbf{1} & 0.978 & 0.929 &
\textbf{1} & 0.947 & 0.890\cr 
    K-ARMA[S] & \textbf{1} & 0.944 & 0.859 & \textbf{1} & 0.903 & 0.794 & \textbf{1} & \textbf{1} & 0.965 & \textbf{1} & \textbf{1} & 0.943\cr
    K-ARMA[M] & \textbf{1} & \textbf{0.965} & \textbf{0.877} & \textbf{1} & \textbf{1} & \textbf{0.816} &
   \textbf{1} & \textbf{1} & \textbf{1} & \textbf{1} & \textbf{1} & \textbf{1} \cr
   SMCE & 0.975 & 0.902 & 0.755 & 0.931 & 0.857 & 0.629 & \textbf{1} & 0.940 & 0.886 & \textbf{1} & 0.899 & 0.828 \cr
    \bottomrule
    \end{tabular}
  \end{threeparttable}
  }
\end{table*}

\begin{figure}
  \centering
  \subfloat[Estimated state label]{
  \includegraphics[width = .7\columnwidth]{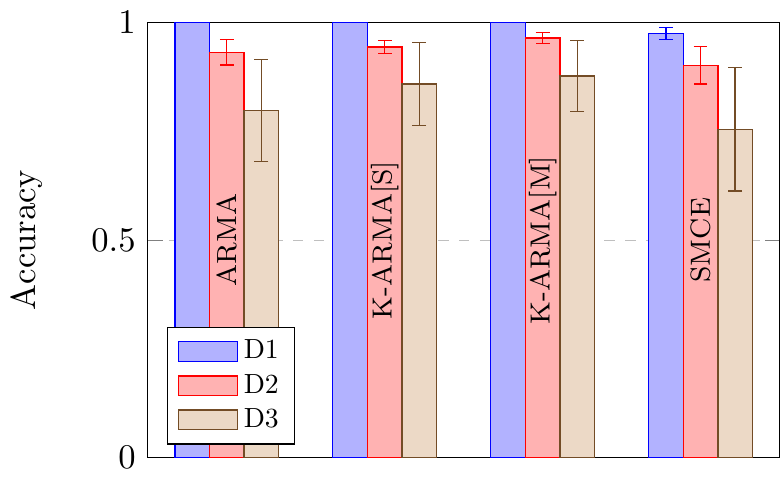}
  }
  
\subfloat[True state label]{
\includegraphics[width = .7\columnwidth]{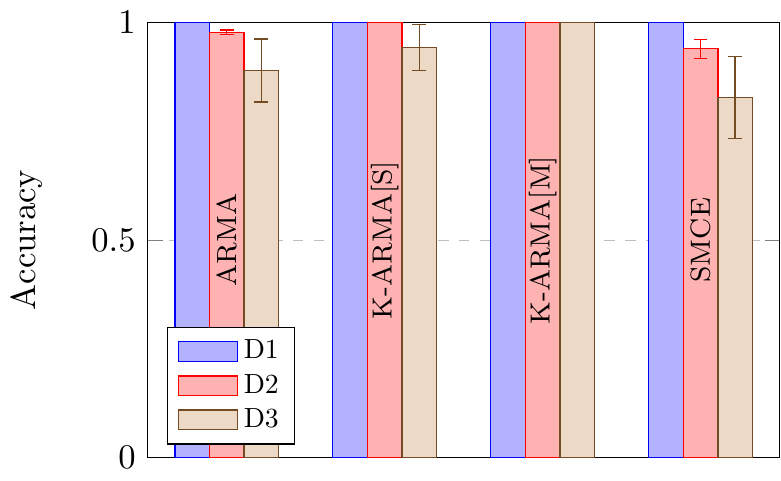}
    }
    \caption{Subnetwork-state-sequence clustering results of synthetic EEG
      datasets. (a) With estimated state label; (b) with true state
      label.} \label{Fig:synthetic.EEG:task tracking}
\end{figure}

Results of subnetwork-state-sequence clustering for the EEG synthetic data are
shown in Table~\ref{Table:synthetic.EEG:task.tracking}. The utilized parameters
and kernel functions were chosen to be the same as in the previous case of
community detection. ARMA, K-ARMA[S] and K-ARMA[M] got 100\% accuracy and NMI
for dataset D1, since the standard deviation of ``noise'' matrices are small and
dataset D1 does not include outliers. It can be clearly verified by
Table~\ref{Table:synthetic.EEG:task.tracking} that if the ``true'' state labels
are known beforehand, and thus step~\ref{Step:Framework:Task:find.states.first}
in Alg.~\ref{Algo:the.clustering.framework} is not needed, then all values of
clustering accuracies and NMI increase. Fig.~\ref{Fig:synthetic.EEG:task
  tracking} depicts also the standard deviations of the results of
Table~\ref{Table:synthetic.EEG:task.tracking}. By comparing the two figures in
Fig.~\ref{Fig:synthetic.EEG:task tracking}, it can be clearly seen that if the
``true'' state labels are known beforehand, \ie,
step~\ref{Step:Framework:Task:find.states.first} in
Alg.~\ref{Algo:the.clustering.framework} is not needed, then all values of
clustering accuracies increase while standard deviations decrease. Similarly to
other tests, K-ARMA[M] shows the highest accuracy on all datasets.

\subsection{Real Data}\label{Sec:real.data}

The open-source real EEG data~\cite{andrzejak2001indications} were used. The
data comprise five sets (A--E), each containing $100$ single-channel segments of
$23.6$-sec duration. The sampling rate of the data was $173.61$ Hz. Only state
clustering is examined, since data~\cite{andrzejak2001indications} do not
contain any connectivity-structure information. Datasets D and E were chosen,
where set D contains only activity measured during seizure free intervals, while
set E contains only seizure activity. The length of the time series extracted
from the data was set equal to $4,096$. Methods ARMA, K-ARMA[S], K-ARMA[M], SMCE
and K-means were validated. 3D-WTA did not perform well on those datasets, and
thus, its results are not shown here.

\begin{figure}
  \centering
  \includegraphics[width = .7\columnwidth]{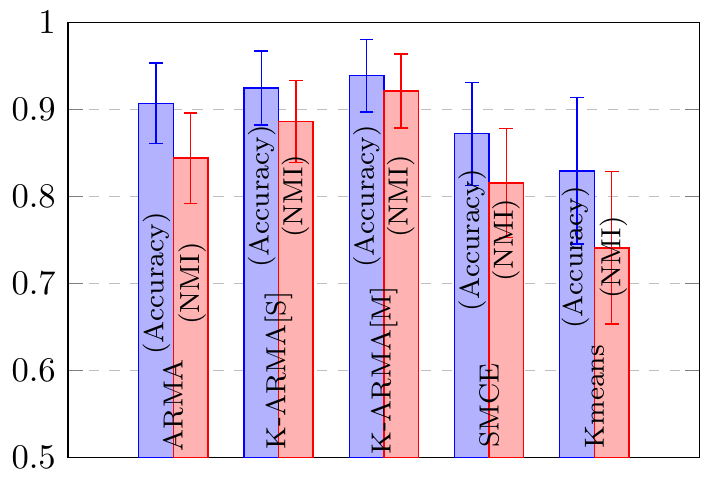}

  \caption{Real EEG data: State clustering.}
  \label{Fig:real.EEG:state.clustering}
\end{figure}

Toward a realistic scenario, the time series of the sets D and E are
concatenated to create a single time series with length $2\cdot 4,096 =
8,192$. The network has 100 nodes, so $|\mathcal{N}| \coloneqq 100$. Parameters
of ARMA, K-ARMA[S] and K-ARMA[M] are defined as: $N \coloneqq 200$,
$m \coloneqq 3$, $\rho \coloneqq 2$, $\tau_f \coloneqq 1800$,
$\tau_b \coloneqq 50$. In K-ARMA[S], the kernel function is set equal to
$\kappa_{\text{G}; 0.6}(\cdot, \cdot)$, while in K-ARMA[M]
$\kappa(\cdot, \cdot) \coloneqq 0.5\, \kappa_{\text{G}; 0.6}(\cdot, \cdot) +
0.5\, \kappa_{\text{L}; 0.8}(\cdot, \cdot)$. Due to the sliding-window
implementation in the proposed framework, there are cases where the sliding
window captures samples from both the D and E time series. The features
extracted from those cases are labeled as cluster
$3$. Fig.~\ref{Fig:real.EEG:state.clustering} depicts also the standard
deviations of the results of Table~\ref{Table:Real.EEG:state.clustering}.
K-ARMA[M] scores the highest clustering-accuracy and NMI values with the
smallest standard deviation. In the case where the time series D and E are not
concatenated, the case of sliding windows capturing data from both clusters
(states) is non-existent. In such a binary-decision case, the confusion-matrix
results of ARMA, K-ARMA[S] and K-ARMA[M] are $\text{TPR} = 1$, $\text{FPR} = 0$,
$\text{FNR} = 0$ and $\text{TNR} = 1$, while for SMCE, $\text{TPR} = 0.952$,
$\text{FPR} = 0.060$, $\text{FNR} = 0.048$ and $\text{TNR} = 0.939$.

\begin{table}[tp]
  \centering
  \caption{Real EEG data: State clustering}
  \label{Table:Real.EEG:state.clustering}
  \begin{tabular}{c|c|c}
    \toprule
    Methods & Clustering accuracy & NMI\cr\midrule
    ARMA & 0.907 & 0.844 \cr
    K-ARMA[S] & 0.925 & 0.886 \cr
    K-ARMA[M] & \textbf{0.939} & \textbf{0.921} \cr
    SMCE & 0.873 & 0.815 \cr
    Kmeans & 0.829 & 0.741 \cr                           
      \bottomrule
  \end{tabular}
\end{table}


Real fMRI behavioral data, acquired from the Stellar Chance 3T scanner (SC3T) at
the University of Pennsylvania, were used to cluster different states. The time
series in data are collected in two arms before and after an inhibitory sequence
of transcranial magnetic stimulation (TMS) known as continuous theta burst
stimulation~\cite{huang2005theta}. Real and Sham stimulation of two different
tasks were applied for TMS. The two behavioral tasks are: 1) Navon task: A big
shape made up of little shapes is shown on the screen. The big shape can either
be green or white in color. If green, participant identifies the big shape,
while if white, the participant identifies the little shape. The task was
presented in three blocks: All white stimuli, all green stimuli, and switching
between colors on 70\% of trials to introduce switching demands. Responses given
via button box are in the order of circle, x, triangle, square; 2) Stroop task:
Words are displayed in different color inks. There are two difficulty
conditions; one where subjects respond to words that introduced low color-word
conflict (far, deal, horse, plenty) or high conflict with color words differing
from the color the word is printed in (\eg red printed in blue, green printed
in yellow, etc.)~\cite{medaglia2018functional}. The participant has to tell the
color of the ink the word is printed in using a button box in the order of red,
green, yellow, blue.

Each BOLD time series was collected during an $8$min scan with
$\text{TR}=500$ms, which means that the length of time series is $956$. The time
series has $83$ cortical and subcortical regions so
$|\mathcal{N}| \coloneqq 83$. To test the state clustering results of fMRI time
series, 3 states are concatenated to create a single time series with length
$3\times 956 = 2,868$. The 3 states are: 1) Before real stimulation of the Navon
task; 2) after real stimulation of the Navon task; and 3) after real stimulation
of the Stroop task.

Parameters of ARMA, K-ARMA[S] and K-ARMA[M] are defined as: $N \coloneqq 180$,
$m \coloneqq 4$, $\rho \coloneqq 2$, $\tau_f \coloneqq 350$,
$\tau_b \coloneqq 20$. In K-ARMA[S], the kernel function is set equal to
$\kappa_{\text{G}; 0.45}(\cdot, \cdot)$, while in K-ARMA[M]
$\kappa(\cdot, \cdot) \coloneqq 0.3\, \kappa_{\text{G}; 0.25}(\cdot, \cdot) +
0.3\, \kappa_{\text{G}; 0.9}(\cdot, \cdot) + 0.4\, \kappa_{\text{L};0.75}(\cdot,
\cdot)$. Notice here that due to the sliding-window implementation in the
proposed framework, there are cases where the sliding window captures samples
from two consecutive states. 

\begin{table}[tp]
  \centering
  \caption{Real fMRI data: State clustering results}
  \label{Table:Real.fMRI:state.clustering}
  \begin{tabular}{c|c|c}
    \toprule
    Methods & Clustering accuracy & NMI\cr\midrule
    ARMA & 0.885 & 0.809 \cr
    K-ARMA[S] & 0.904 & 0.843 \cr
    K-ARMA[M] & \textbf{0.919} & \textbf{0.875} \cr
    SMCE & 0.893 & 0.816 \cr
    Kmeans & 0.801 & 0.720 \cr                           
      \bottomrule
  \end{tabular}
\end{table}

\begin{figure}[htpb!]
  \centering
  \includegraphics[width = .7\columnwidth]{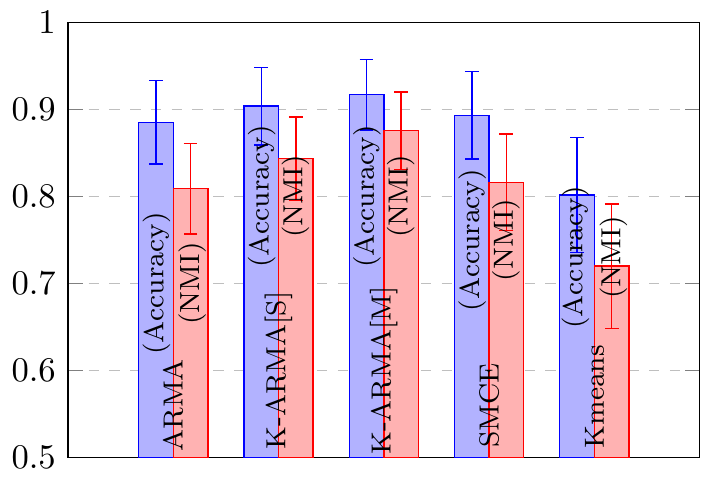}

  \caption{Real fMRI data: State clustering.}
  \label{Fig:real.fMRI:state.clustering}
\end{figure}

Results of state clustering on real fMRI data are revealed in
Table~\ref{Table:Real.fMRI:state.clustering}. Fig.~\ref{Fig:real.fMRI:state.clustering}
depicts also the standard deviations of the results of
Table~\ref{Table:Real.fMRI:state.clustering}. Again, K-ARMA[M] exhibits the
highest clustering-accuracy and NMI values, with the smallest standard deviation
among all employed methods.


\section{Conclusions}\label{Sec:conclusions}

This paper introduced a novel clustering framework to perform all possible
clustering tasks in dynamic (brain) networks: state clustering, community
detection and subnetwork-state-sequence tracking/identification. Features were
extracted by a kernel-based ARMA model, with column spaces of observability
matrices mapped to the Grassmann manifold (Grassmannian). A clustering
algorithm, the geodesic clustering with tangent spaces, was also provided to
exploit the rich underlying Riemannian geometry of the Grassmannian. The
framework was validated on multiple simulated and real datasets and compared
against state-of-the-art clustering algorithms. Test results demonstrate that
the proposed framework outperforms the competing methods in clustering states,
identifying community structures, and tracking multiple subnetwork state sequences
which may span several network states. Current research effort includes finding
ways to reduce the size of the computational footprint of the framework, and
techniques to reject network-wide outlier data.

\appendices

\section{Reproducing Kernel Hilbert Spaces}\label{App:RKHS}

A reproducing kernel Hilbert space $\mathcal{H}$, equipped with inner product
$\innerp{\cdot}{\cdot}_{\mathcal{H}}$, is a functional space where each point
$g\in \mathcal{H}$ is a function
$g: \Real^q\to \Real: \vect{y} \mapsto g(\vect{y})$, for some $q\in\IntegerPP$,
s.t.\ the mapping $g\mapsto g(\vect{y})$ is continuous, for any choice of
$\vect{y}$~\cite{Aronszajn.50, Scholkopf.Smola.Book,
  Slavakis.OL.BookChapter.14}. There exists a kernel function
$\kappa(\cdot, \cdot): \Real^q\times \Real^q \to \Real$, unique to
$\mathcal{H}$, such that (s.t.)
$\varphi(\vect{y}) \coloneqq \kappa(\vect{y}, \cdot)\in \mathcal{H}$ and
$g(\vect{y}) = \innerp{g}{\varphi(\vect{y})}_{\mathcal{H}}$, for any
$g\in \mathcal{H}$ and any $\vect{y}\in \Real^q$~\cite{Aronszajn.50,
  Slavakis.OL.BookChapter.14}. The latter property is the reason for calling
kernel $\kappa$ reproducing, and yields the celebrated ``kernel trick'':
$\kappa(\vect{y}_1, \vect{y}_2) = \innerp{\kappa(\vect{y}_1,
  \cdot)}{\kappa(\vect{y}_2, \cdot)}_{\mathcal{H}} =
\innerp{\varphi(\vect{y}_1)}{\varphi(\vect{y}_2)}_{\mathcal{H}}$, for any
$\vect{y}_1, \vect{y}_2 \in \Real^q$.

Popular examples of reproducing kernels are:
\begin{enumerate}

\item The linear
  $\kappa_{\text{lin}}(\vect{y}_1, \vect{y}_2) \coloneqq \vect{y}_1^{\intercal}
  \vect{y}_2$, where space $\mathcal{H}$ is nothing but $\Real^q$;

\item the Gaussian
  $\kappa_{\text{G}; \sigma}(\vect{y}_1, \vect{y}_2) \coloneqq
  \exp[-\norm{\vect{y}_1 - \vect{y}_2}^2 / (2\sigma^2)]$, where
  $\sigma\in\RealPP$ and $\norm{\cdot}$ is the standard Euclidean norm. In this
  case, $\mathcal{H}$ is infinite dimensional~\cite{Slavakis.OL.BookChapter.14};

\item the Laplacian
  $\kappa_{\text{L}; \sigma}(\vect{y}_1, \vect{y}_2) \coloneqq \exp[-\norm{\vect{y}_1 -
    \vect{y}_2}_1 / \sigma]$, where $\norm{\cdot}_1$ stands for the
  $\ell_1$-norm~\cite{Sriperumbudur.10}; and

\item the polynomial
  $\kappa_{\text{poly}; r}(\vect{y}_1, \vect{y}_2) \coloneqq
  (\vect{y}_1^{\intercal} \vect{y}_2 + 1)^r$, for some parameter
  $r\in\IntegerPP$.

\end{enumerate}
There are several ways of generating reproducing kernels via certain operations
on well-known kernel functions such as convex combinations, products,
\etc~\cite{Scholkopf.Smola.Book}.

Define $\mathcal{H}^p$, for some $p\in \IntegerPP$, as the space whose points
take the following form:
$\bm{g} \coloneqq [g_1, \ldots, g_p]^{\intercal} \in \mathcal{H}^p$ s.t.\
$g_j\in\mathcal{H}$, $\forall j\in\overline{1,p}$, where $\overline{1,p}$ is a
compact notation for $\{1, \ldots, p\}$. For $p'\in \IntegerPP$ and given a
matrix $\vect{A}\coloneqq [a_{ij}] \in \Real^{p'\times p}$, the product
$\vect{A}\bm{g}\in \mathcal{H}^{p'}$ stands for the vector-valued function whose
$i$th entry is $\sum_{j=1}^p a_{ij}g_j$. Similarly, define
$\mathcal{H}^{p_1 \times p_2}$, for some $p_1, p_2\in \IntegerPP$, as the space
comprising all
\begin{align*}
  \bm{\mathcal{G}} \coloneqq
  \begin{bmatrix}
    g_{11} & \cdots & g_{1p_2} \\
    \vdots & \ddots & \vdots \\
    g_{p_11} & \cdots & g_{p_1p_2}
  \end{bmatrix}
  \in \mathcal{H}^{p_1 \times p_2} \,,
\end{align*}
s.t.\ $g_{ij}\in\mathcal{H}$, $\forall i\in\overline{1,p_1}$,
$\forall j\in\overline{1,p_2}$. Moreover, given
$\bm{\mathcal{G}}\in \mathcal{H}^{p_1 \times p}$ and
$\bm{\mathcal{G}}' \in \mathcal{H}^{p \times p_2}$, define the ``product''
$\bm{\mathcal{G}} \Kprod \bm{\mathcal{G}}'$ as the $p_1\times p_2$ matrix whose
$(i,j)$th entry is
\begin{align*}
  [\bm{\mathcal{G}} \Kprod \bm{\mathcal{G}}']_{ij} \coloneqq \sum\nolimits_{l=1}^p
  \innerp{g_{il}}{g_{lj}'}_{\mathcal{H}} \,.
\end{align*}
In the case where $g_{il} \coloneqq \varphi(\vect{y}_{il}) = \kappa(\vect{y}_{il}, \cdot)$ and
$g_{lj}' \coloneqq \varphi(\vect{y}_{lj}') = \kappa(\vect{y}_{lj}', \cdot)$, for some
$\vect{y}_{il}, \vect{y}_{lj}'$, as in \eqref{low.rank.formula}, then the kernel trick suggests that
the previous formula simplifies to
$[\bm{\mathcal{G}} \Kprod \bm{\mathcal{G}}']_{ij} = \sum_{l=1}^p \kappa(\vect{y}_{il},
\vect{y}_{lj}')$.

\section{Proof of Proposition~\ref{Prop:O}}\label{App:prove.prop}

By considering a probability space $(\Omega, \Sigma, \Prob)$, a basis
$\Set{e_n}_{n\in\IntegerPP}$ of $\mathcal{H}$, and by omitting most of the
entailing measure-theoretic details, the expectation of
$g = \sum_{n\in\IntegerPP} \gamma_ne_n \in \mathcal{H}$, where
$\Set{\gamma_n}_{n\in\IntegerPP}$ are real-valued RVs, is defined as
$\Expect(g) \coloneqq \sum_{n\in\IntegerPP} \Expect(\gamma_n)e_n$, provided that
the latter sum converges in $\mathcal{H}$. Conditional expectations are
similarly defined. All of the expectations appearing in this manuscript are
assumed to exist. Due to the linearity of the inner product
$\innerp{\cdot}{\cdot}_{\mathcal{H}}$, it can be verified that the conditional
expectation
$\Expect\{\innerp{g}{g'}_{\mathcal{H}} \given g'\} = \Expect\{ \sum_{n, n'}
\gamma_n \gamma_{n'} \innerp{e_n}{e_{n'}}_{\mathcal{H}} \given g' \} = \sum_{n'}
\gamma_{n'} \sum_n \Expect\{ \gamma_n \given g' \}
\innerp{e_n}{e_{n'}}_{\mathcal{H}} = \innerp{\sum_n \Expect\{ \gamma_n \given g'
  \} e_n}{ \sum_{n'} \gamma_{n'}e_{n'}}_{\mathcal{H}} = \innerp{\Expect\{g\given
  g'\}}{g'}_{\mathcal{H}}$, and
$\Expect\{\innerp{g}{g'}_{\mathcal{H}}\} =
\innerp{\Expect(g)}{\Expect(g')}_{\mathcal{H}}$ in the case where $g$ and $g'$
are independent. It can be similarly verified that these properties, which hold
for the inner product $\innerp{\cdot}{\cdot}$, are inherited by $\Kprod$.

Induction on \eqref{KARMA} suggests that $\forall \tau\in\IntegerP$,
$\bm{\varphi}_{t+\tau} = \vect{CA}^{\tau} \bm{\psi}_t + \sum_{k=1}^{\tau}
\vect{CA}^{\tau-k} \bm{\omega}_{t+k} + \bm{\upsilon}_{t+\tau}$, where
$\sum_{k=1}^{0} \vect{CA}^{-k} \bm{\omega}_{t+k} \coloneqq \vect{0}$. Then,
\begin{align}
  \bm{f}_t
  & \coloneqq \left[ \bm{\varphi}_t^{\intercal}, \bm{\varphi}_{t+1}^{\intercal},
    \ldots, \bm{\varphi}_{t+m-1}^{\intercal} \right]^{\intercal} \notag \\
  & = \vect{O} \bm{\psi}_t + \bm{e}_t \,, \label{phi.O}
\end{align}
where
\begin{align*}
  \bm{e}_t \coloneqq \begin{bmatrix}
    \bm{\upsilon}_t \\
    \vect{C} \bm{\omega}_{t+1} + \bm{\upsilon}_{t+1} \\
    \sum_{k=1}^2 \vect{CA}^{2-k} \bm{\omega}_{t+k} + \bm{\upsilon}_{t+2} \\
    \vdots \\
    \sum_{k=1}^{m-1} \vect{CA}^{m-1-k} \bm{\omega}_{t+k} + \bm{\upsilon}_{t+m-1}
  \end{bmatrix} \in \mathcal{H}^{mN} \,.
\end{align*}

By observing that
$\bm{\mathcal{F}}_t = [\bm{f}_t, \bm{f}_{t+1}, \ldots, \bm{f}_{t+\tau_{\text{f}}
  -1}]$, it can be verified that
\begin{align*}
  \bm{\mathcal{F}}_t = \vect{O} \left[\bm{\psi}_t, \bm{\psi}_{t+1}, \ldots,
  \bm{\psi}_{t+\tau_{\text{f}} -1} \right] + \left[\bm{e}_t, \bm{e}_{t+1}, \ldots,
  \bm{e}_{t+\tau_{\text{f}} -1} \right] \,.
\end{align*}
Moreover, notice that $\bm{\mathcal{B}}_t = [\bm{b}_t, \bm{b}_{t+1}, \ldots,
\bm{b}_{t+\tau_{\text{f}} -1}]$, where
\begin{align*}
  \bm{b}_t \coloneqq \left[ \bm{\varphi}_t^{\intercal},
  \bm{\varphi}_{t-1}^{\intercal}, \ldots,
  \bm{\varphi}_{t-\tau_{\text{b}}+1}^{\intercal} \right]^{\intercal}
  \in\mathcal{H}^{\tau_{\text{b}}N} \,. 
\end{align*}
Hence,
\begin{alignat*}{2}
  &&& \hspace{-30pt} \tfrac{1}{\tau_{\text{f}}}
  \bm{\mathcal{F}}_{t+1} \Kprod \bm{\mathcal{B}}_t^{\intercal} \\
  & {} = {} && \tfrac{1}{\tau_{\text{f}}} \vect{O} \left[\bm{\psi}_{t+1},
    \ldots, \bm{\psi}_{t+\tau_{\text{f}}} \right] \Kprod
  \bm{\mathcal{B}}_t^{\intercal} \\ 
  &&& + \tfrac{1}{\tau_{\text{f}}} \left[\bm{e}_{t+1}, \ldots,
    \bm{e}_{t+\tau_{\text{f}}} \right] \Kprod \bm{\mathcal{B}}_t^{\intercal} \\
  & = && \vect{O}\, \tfrac{1}{\tau_{\text{f}}} \sum\nolimits_{l=1}^{\tau_{\text{f}}}
  \bm{\psi}_{t+l} \Kprod \bm{b}_{t+l-1}^{\intercal} \\
  &&& + \tfrac{1}{\tau_{\text{f}}} \sum\nolimits_{l=1}^{\tau_{\text{f}}}
  \bm{e}_{t+l} \Kprod \bm{b}_{t+l-1}^{\intercal} \\
  & = && \vect{O}\, \tfrac{1}{\tau_{\text{f}}} \sum\nolimits_{l=1}^{\tau_{\text{f}}}
  \bm{\psi}_{t+l} \Kprod [\bm{\psi}_{t+l-1}^{\intercal} \vect{C}^{\intercal},
  \ldots, \bm{\psi}_{t+l-\tau_{\text{b}}}^{\intercal} \vect{C}^{\intercal}] \\
  &&& + \vect{O}\, \tfrac{1}{\tau_{\text{f}}} \sum\nolimits_{l=1}^{\tau_{\text{f}}}
  \bm{\psi}_{t+l} \Kprod [\bm{\upsilon}_{t+l-1}^{\intercal}, \ldots,
  \bm{\upsilon}_{t+l-\tau_{\text{b}}}^{\intercal}] \\
  &&& + \tfrac{1}{\tau_{\text{f}}} \sum\nolimits_{l=1}^{\tau_{\text{f}}}
  \bm{e}_{t+l} \Kprod \bm{b}_{t+l-1}^{\intercal} \,,
\end{alignat*}
and \eqref{low.rank.formula} is established under the following definitions:
\begin{alignat}{2}
  \bm{\Pi}_{t+1}
  & {} \coloneqq {} &&  \tfrac{1}{\tau_{\text{f}}}
  \sum\nolimits_{l=1}^{\tau_{\text{f}}}
  \bm{\psi}_{t+l} \Kprod [\bm{\psi}_{t+l-1}^{\intercal} \vect{C}^{\intercal},
  \ldots, \bm{\psi}_{t+l-\tau_{\text{b}}}^{\intercal} \vect{C}^{\intercal}] \,,
  \notag \\
  \bm{\mathcal{E}}_{t+1}^{\tau_{\text{f}}} 
  & \coloneqq && \vect{O}\, \tfrac{1}{\tau_{\text{f}}}
  \sum\nolimits_{l=1}^{\tau_{\text{f}}} \bm{\psi}_{t+l} \Kprod
  [\bm{\upsilon}_{t+l-1}^{\intercal}, \ldots, 
  \bm{\upsilon}_{t+l-\tau_{\text{b}}}^{\intercal}] \notag \\
  &&& + \tfrac{1}{\tau_{\text{f}}} \sum\nolimits_{l=1}^{\tau_{\text{f}}}
  \bm{e}_{t+l} \Kprod \bm{b}_{t+l-1}^{\intercal} \,. \label{E}
\end{alignat}

By virtue of the independency between $(\bm{\psi}_t)_t$ and
$(\bm{\upsilon}_t)_t$, the zero-mean assumption on $(\bm{\upsilon}_t)_t$, as
well as standard properties of the conditional
expectation~\cite[\S9.7(k)]{Williams.book} with respect to independency, it can
be verified that
\begin{align}
  & \Expect\{ \bm{\psi}_{t+l} \Kprod
  [\bm{\upsilon}_{t+l-1}^{\intercal}, \ldots, 
  \bm{\upsilon}_{t+l-\tau_{\text{b}}}^{\intercal}] \given \bm{\psi}_{t+l} \}
    \notag \\
  & = \bm{\psi}_{t+l} \Kprod [ \Expect\{\bm{\upsilon}_{t+l-1}^{\intercal}\}, \ldots, 
  \Expect\{\bm{\upsilon}_{t+l-\tau_{\text{b}}}^{\intercal}\}] = \vect{0}
    \,. \label{E1}
\end{align}
Moreover, for any $i\in\overline{1,m}$ and any
$j\in\overline{1,\tau_{\text{b}}}$, the $(i,j)$th $N\times N$ block of the
second term in the expression of $\bm{\mathcal{E}}_{t+1}^{\tau_{\text{f}}}$ in
\eqref{E} becomes equal to
\begin{align}
  & \sum\nolimits_{k=1}^{i-1} \vect{CA}^{i-1-k}
    \tfrac{1}{\tau_{\text{f}}} \sum\nolimits_{l=1}^{\tau_{\text{f}}}
    \bm{\omega}_{t+l+k} \Kprod \bm{\varphi}_{t+l-j}^{\intercal} \notag \\
  & \hphantom{=\ } + \tfrac{1}{\tau_{\text{f}}} \sum\nolimits_{l=1}^{\tau_{\text{f}}}
    \bm{\upsilon}_{t+l+i-1} \Kprod \bm{\varphi}_{t+l-j}^{\intercal} \notag \\
  & = \sum\nolimits_{k=1}^{i-1} \vect{CA}^{i-1-k}
    \tfrac{1}{\tau_{\text{f}}} \sum\nolimits_{l=1}^{\tau_{\text{f}}}
    \bm{\omega}_{t+l+k} \Kprod \bm{\psi}_{t+l-j}^{\intercal}
    \vect{C}^{\intercal} \notag \\
  & \hphantom{=\ } + \sum\nolimits_{k=1}^{i-1} \vect{CA}^{i-1-k}
    \tfrac{1}{\tau_{\text{f}}} \sum\nolimits_{l=1}^{\tau_{\text{f}}}
    \bm{\omega}_{t+l+k} \Kprod \bm{\upsilon}_{t+l-j}^{\intercal} \notag \\  
  & \hphantom{=\ } + \tfrac{1}{\tau_{\text{f}}}
    \sum\nolimits_{l=1}^{\tau_{\text{f}}} \bm{\upsilon}_{t+l+i-1} \Kprod
    \bm{\psi}_{t+l-j}^{\intercal} \vect{C}^{\intercal} \notag \\
  & \hphantom{=\ } + \tfrac{1}{\tau_{\text{f}}}
    \sum\nolimits_{l=1}^{\tau_{\text{f}}} \bm{\upsilon}_{t+l+i-1} \Kprod
    \bm{\upsilon}_{t+l-j}^{\intercal}  \,. \label{block.E2}
\end{align}
Since $t+l+k > t+l > t+l-j$ and $t+l+i-1 \geq t+l > t+l-j$, $\bm{\psi}_{t+l-j}$
precedes $\bm{\omega}_{t+l+k}$ on the time axis, while $\bm{\upsilon}_{t+l+i-1}$
precedes $\bm{\upsilon}_{t+l-j}$. Hence, due to independency,
$\Expect\{ \bm{\omega}_{t+l+k} \Kprod \bm{\psi}_{t+l-j}^{\intercal} \given
\bm{\psi}_{t'} \} = \Expect\{ \bm{\omega}_{t+l+k} \given \bm{\psi}_{t'} \}
\Kprod \bm{\psi}_{t+l-j}^{\intercal} = \Expect\{ \bm{\omega}_{t+l+k} \} \Kprod
\bm{\psi}_{t+l-j}^{\intercal} = \vect{0}$, and
$\Expect \{ \bm{\upsilon}_{t+l+i-1} \Kprod \bm{\upsilon}_{t+l-j}^{\intercal}
\given \bm{\psi}_{t'} \} = \Expect \{ \bm{\upsilon}_{t+l+i-1}\} \Kprod \Expect\{
\bm{\upsilon}_{t+l-j}^{\intercal} \} = \vect{0}$. It can be also similarly
verified that
$\Expect\{ \bm{\omega}_{t+l+k} \Kprod \bm{\upsilon}_{t+l-j}^{\intercal} \given
\bm{\psi}_{t'} \} = \vect{0}$ and
$\Expect\{ \bm{\upsilon}_{t+l+i-1} \Kprod \bm{\psi}_{t+l-j}^{\intercal} \given
\bm{\psi}_{t'} \} = \vect{0}$. As a result, the conditional expectation of
\eqref{block.E2}, given $\bm{\psi}_{t'}$, becomes $\vect{0}$. This observation
and \eqref{E1} establish claim \eqref{cond.expectation} of the proposition.

Under the assumptions on wide-sense stationarity, the covariance sequences of
the processes $(\bm{\omega}_t \Kprod \bm{\psi}_{t-\tau}^{\intercal})_t$,
$(\bm{\omega}_t \Kprod \bm{\upsilon}_{t-\tau}^{\intercal})_t$,
$(\bm{\upsilon}_t \Kprod \bm{\psi}_{t-\tau}^{\intercal})_t$,
$( \bm{\psi}_t \Kprod \bm{\upsilon}_{t-\tau}^{\intercal})_t$,
$(\bm{\upsilon}_t \Kprod \bm{\upsilon}_{t-\tau}^{\intercal})_t$,
$\forall\tau \in\IntegerPP$, are summable over all lags; in fact, the
covariances of non-zero lags become zero due to the assumptions on
independency. Hence, by the mean-square ergodic theorem~\cite{Petersen}, sample
averages of the previous processes converge in the mean-square
($\mathcal{L}_2$-) sense to their ensemble means. For example, applying
$\lim_{\tau_{\text{f}} \to\infty}$, in the mean-square sense, to the first part
of $\bm{\mathcal{E}}_{t+1}^{\tau_{\text{f}}}$ in \eqref{E} and by
recalling standard properties of the conditional
expectation~\cite[\S9.7(a)]{Williams.book} yield
\begin{align}
  & \vect{O} \lim_{\tau_{\text{f}} \to\infty} \tfrac{1}{\tau_{\text{f}}}
  \sum\nolimits_{l=1}^{\tau_{\text{f}}} \bm{\psi}_{t+l} \Kprod
  [\bm{\upsilon}_{t+l-1}^{\intercal}, \ldots, 
    \bm{\upsilon}_{t+l-\tau_{\text{b}}}^{\intercal}] \notag \\
  & = \vect{O} \Expect\{ \bm{\psi}_{t+l} \Kprod
    [\bm{\upsilon}_{t+l-1}^{\intercal}, \ldots, 
    \bm{\upsilon}_{t+l-\tau_{\text{b}}}^{\intercal}] \} \notag \\
  & = \vect{O} \Expect\{ \Expect\{ \bm{\psi}_{t+l} \Kprod
  [\bm{\upsilon}_{t+l-1}^{\intercal}, \ldots, 
  \bm{\upsilon}_{t+l-\tau_{\text{b}}}^{\intercal}] \given \bm{\psi}_{t+l} \} \}
    = \vect{0} \,. \label{E1.ergodicity}
\end{align}
By following similar arguments, it can be verified that the application of
$\lim_{\tau_{\text{f}} \to\infty}$ to \eqref{block.E2} renders the second part
of \eqref{E} equal to $\vect{0}$. This finding and \eqref{E1.ergodicity}
establish the final claim of the proposition.



\end{document}